\documentclass[lettersize,journal]{IEEEtran}

\IEEEoverridecommandlockouts                              

\usepackage{amsmath,amsfonts}
\usepackage{algorithmic}
\usepackage{algorithm}
\usepackage{array}
\usepackage[caption=false,font=normalsize,labelfont=sf,textfont=sf]{subfig}
\usepackage{textcomp}
\usepackage{stfloats}
\usepackage{url}
\usepackage{verbatim}
\usepackage{graphicx}
\usepackage{cite}
\usepackage{xcolor}  
\usepackage{graphics} 
\usepackage{graphicx}
\usepackage[utf8]{inputenc}
\usepackage{amssymb}
\usepackage{amsmath} 
\usepackage{multirow}
\usepackage{booktabs}
\usepackage{hyperref}
\usepackage{cite}
\usepackage{array}
\usepackage{pifont}
\usepackage{bbding}
\usepackage{makecell}
\usepackage[normalem]{ulem}

\usepackage{tabularx}
\usepackage{booktabs}
\hypersetup{hypertex=true,
colorlinks=true,
linkcolor=blue,
anchorcolor=blue,
citecolor=blue}

\usepackage{geometry}
\title{On-the-fly Feedback SfM: Online Explore-and-Exploit UAV Photogrammetry with Incremental Mesh Quality-Aware Indicator and Predictive Path Planning}
\author{Liyuan Lou$^{\dag}$, Wanyun Li$^{\dag}$, Wentian Gan$^{\dag}$, Yifei Yu, Tengfei Wang, Xin Wang,~\IEEEmembership{Member,~IEEE}, Zongqian Zhan,~\IEEEmembership{Member,~IEEE}

\thanks{$^{\dag}$Liyuan Lou, Wanyun Li and Wentian Gan contributed equally to this work.}
\thanks{This work was supported by the National Natural Science Foundation of China (No. 42301507) and the Luojia Undergraduate Innovation Research Fund of Wuhan University. (\textit{Corresponding author: Xin Wang and Zongqian Zhan, xwang@sgg.whu.edu.cn, zqzhan@sgg.whu.edu.cn})}

\thanks{All the authors are with the School of Geodesy and Geomatics, Wuhan University, Wuhan, 430079, China.

}

}

\begin{document}
\newcounter{tmp} 
\maketitle

\begin{abstract}
Compared with conventional offline UAV photogrammetry, real-time UAV photogrammetry is essential for time-critical geospatial applications such as disaster response and active digital-twin maintenance. However, most existing methods focus on processing captured images or sequential frames in real time, without explicitly evaluating the quality of the on-the-go 3D reconstruction or providing guided feedback to enhance image acquisition in the target area. This work presents On-the-fly Feedback SfM, an \textit{explore-and-exploit} framework for real-time UAV photogrammetry, enabling iterative exploration of unseen regions and exploitation of already observed and reconstructed areas in near real time. Built upon SfM on-the-fly\cite{zhan2025sfm}$\footnote{See more at \href{ https://yifeiyu225.github.io/on-the-flySfMv2.github.io/}{ https://yifeiyu225.github.io/on-the-flySfMv2.github.io/}}$, the proposed method integrates three modules: (1) online incremental coarse-mesh generation for dynamically expanding sparse 3D point cloud; (2) online mesh quality assessment with actionable indicators; and (3) predictive path planning for on-the-fly trajectory refinement. Comprehensive experiments demonstrate that our method achieves in-situ reconstruction and evaluation in near real time while providing actionable feedback that markedly reduces coverage gaps and re-flight costs. Via the integration of data collection, processing, 3D reconstruction and assessment, and online feedback, our on-the-fly feedback SfM could be an alternative for the transition from traditional passive working mode to a more intelligent and adaptive exploration workflow. Code is now available at~\href{https://github.com/IRIS-LAB-whu/OntheflySfMFeedback}{https://github.com/IRIS-LAB-whu/OntheflySfMFeedback}.
\end{abstract}

\section{Introduction}

In recent decades, traditional Unmanned Aerial Vehicle (UAV) photogrammetry has been widely studied and applied in an offline manner, i.e., "Process after Capture",  while interest in real-time performance has increasingly grown for applications such as rapid disaster response, active digital twin construction, etc. However, most existing real-time works take video frames or images with a well planned flight\cite{xiao2025rto,xiao2025novel},   which may perpetuate degenerated data for 3D reconstruction; for example, incomplete coverage is often found after flight, necessitating costly re-surveys\cite{nex2014uav, colomina2014unmanned}. Recently, UAVs are no longer merely passive imaging platforms \cite{zhang2021fusion} for collecting data and are more commonly used as \textit{autonomous explorers} for 3D reconstruction, capable of optimizing data capture in real time\cite{mostegel2016uav, song2021view}. The key challenges include not only processing captured images on the fly or in real time, but also indicating the quality and completeness of the 3D reconstruction online and timely offering feedback or guidance on optimal viewpoints to the UAV during the mission itself.

In the past few years, optimized view planning for UAV photogrammetry, in an explore-then-exploit manner, has been extensively studied to optimize flying path  \cite{zhou2020offsite,maboudi2023review}. The core objective is to compute a set of optimal viewpoints from which images should be captured, such that high-quality 3D reconstructions can be achieved with minimal image acquisition. Typically, a coarse 3D model generated by satellite images is required as priors for planning the viewpoints, and all the subsequent processes are executed offline. To achieve real-time performance, Hoppe et al.\cite{hoppe2012online} incorporated 3D model feedback into online SfM without explicit guidance for image acquisition, while Huang et al.\cite{huang2018active} developed an iterative linear method to efficiently solve multi-view stereo (MVS) and plan the Next-Best-View (NBV). However, these approaches were only evaluated on small-scale toy examples and proved ineffective for large-scale scenes captured using high-resolution UAV imagery. 

This paper presents \textbf{On-the-fly Feedback SfM}, a real-time UAV photogrammetry framework that combines the data acquisition and 3D reconstruction via integration of online SfM, incremental meshing, quality assessment, and predictive path planning. The framework reformulates UAV image capture in an \textit{explore–and–exploit} manner: as each new image arrives, \textit{SfM on-the-fly}\cite{zhan2025sfm} updates camera poses and the evolving sparse point cloud; an online incremental meshing module is employed to yield a coarse but interpretable surface representation, based on which the reconstruction quality assessment is performed by estimating several indicators (e.g., ground sampling distance, redundancy); finally, predictive path planning is determined via investigating these assessments and updates the on-the-fly trajectory, ensuring adaptive exploration of unexplored regions and further refinement of the reconstruction during the mission.

The key contributions of this work are summarized as follows:


\begin{enumerate}
    \item \textbf{Online explore–and-exploit framework for UAV photogrammetry.} 
    To the best of our knowledge, we propose one of the first online frameworks that couples image acquisition, incremental reconstruction, quality assessment, and optimal path planning into a unified \textit{explore–and–exploit} workflow. This enables UAVs not only to passively capture data, but to actively adapt their flight path during missions to improve reconstruction quality in real time.

    \item \textbf{Incremental meshing with mesh quality-aware indicators.} 
    Building upon on-the-fly SfM, our method introduces a dynamic energy function and incremental ray tracing with dynamic graph-cuts to efficiently reconstruct surfaces from growing point cloud.  This mesh serves as the foundation for computing interpretable quality indicators such as ground sampling distance, redundancy, and reprojection error.

    \item \textbf{Predictive path planning via real-time and quality-aware trajectory optimization.}
    A quality-driven path planning module is proposed, featuring a novel pipeline that first detects coverage gaps using the ensemble indicator $Q_{total}$ and DBSCAN clustering, then generates a sparse yet effective set of viewpoints through a multi-constraint viewpoint generation strategy, and finally optimizes the trajectory via an altitude-aware cost function combined with 2-opt refinement\cite{croes1958method}. This enables UAV trajectories to be adaptively updated on the fly, minimizing re-flights while maximizing reconstruction quality.

\end{enumerate}

\section{Related Work}
In this section, some studies most relevant to the proposed online UAV photogrammetric framework are reviewed, including incremental surface reconstruction, feedback mechanisms for image acquisition and reconstruction, active reconstruction and path planning strategies.

\subsection{Incremental Surface Reconstruction}
The conventional solution underlying most existing methods for extracting surfaces from sparse Structure-from-Motion (SfM) point cloud is to perform a Delaunay triangulation (DT) and construct a volumetric tetrahedral mesh, which is then partitioned into free and occupied space through visibility-based optimization. The boundary separating the free and occupied space constitutes the final extracted surface\cite{labatut2009robust}. While highly robust under batch processing scenarios, these traditional methods are commonly designed for offline reconstruction, assuming full access to the complete point cloud. When new points are added, both the Delaunay triangulation and the associated graph-cut optimization must be recomputed globally, leading to significant computational overhead and  making them infeasible for incremental and online workflows, as their computational cost scales sharply with the increasing number of points.

To overcome the above-mentioned limitations, several early works have explored incremental surface extraction from continuously expanding sparse point cloud. The ProFORMA proposed by Pan et al.\cite{pan2009proforma} achieved rapid online model generation by performing Delaunay tetrahedralization on an online SfM point cloud and introduced a novel and efficient probabilistic recursive tetrahedron carving algorithm.
To address potential topological errors during the incremental reconstruction, Yu and Lhuillier\cite{yu2012incremental} proposed an incremental region-growing approach that operated within a continuously updated Delaunay tetrahedralization, it could generate and maintain a hole-free and non-self-intersecting 2-manifold surface. Hoppe et al. \cite{hoppe2013incremental} attempted to adapt the robust but offline graph-cut framework\cite{labatut2009robust} into an incremental manner by introducing two key innovations: a new efficient energy function that relies only on local visibility, and the use of dynamic graph cuts to efficiently update the previous solution rather than re-solving from scratch. This combination allows for near real-time performance that is largely independent of the overall scene size, making it highly suitable for online applications, such as SLAM.

Despite notable algorithmic advancements, existing incremental reconstruction methods have been predominantly validated in ground-based SLAM or a controlled indoor environment. Their scalability and robustness under the demanding conditions of large-scale aerial photogrammetry—characterized by extensive coverage, high-resolution imagery, and frequent inter-occlusions—remain largely unexplored. This gap motivates the development of a real-time, mesh-quality aware reconstruction framework specifically designed for UAV-based applications.

\subsection{Feedback Mechanisms for Image Acquisition and Reconstruction}
To improve image acquisition without quality loss in reconstruction, feedback mechanism is of vital importance with extensive studies. Early attempts to improve 3D reconstruction quality during image acquisition mainly relied on offline feedback mechanisms. 
In this context, an initial flight is conducted to capture imagery following a pre-defined plan, and reconstruction is carried out offline to assess coverage completeness and identify missing or low-quality regions, often by analyzing the density and geometric consistency of the preliminary point cloud or mesh \cite{li2023optimized, nex2014uav,zhang2024guided}. 
Subsequent re-flights are then scheduled to fill in these gaps, a common practice in large-scale mapping to ensure model integrity \cite{james2014mitigating}.
Common approaches include iterative \textit{plan–capture–analyze–replan} pipelines, where coarse 3D models or ortho-mosaics are generated after the first flight, followed by manual or semi-automatic selection of additional viewpoints. 
While this is effective in improving reconstruction completeness, the corresponding offline feedback strategies are inherently time and effort-consuming and unsuitable for time-critical applications such as emergency mapping or rapid site inspection.

In the studies of real-time or online reconstruction, endeavors have been  investigated in online feedback mechanisms that integrate image acquisition and reconstruction into a complete solution. Hoppe et al.~\cite{hoppe2012online} pioneered an online feedback method for SfM, where a sparse point cloud is incrementally updated as new images are captured, and reconstruction completeness is estimated in real time. However, they provided no explicit guidance for viewpoint selection, instead relying on the manual operator to interpret the feedback. Mostegel et al.~\cite{mostegel2016uav} further proposed an autonomous image acquisition framework that predicts the confidence of dense multi-view stereo (MVS) reconstruction on-the-fly, enabling on-site quality assurance. Similarly, Torresani et al.~\cite{torresani2021v} highlighted that real-time feedback during acquisition and processing enables users to actively improve the final reconstruction, preventing missing parts and weak camera networks. These works collectively demonstrate the potential of completing the acquisition–reconstruction loop, while also motivating the requirement for automated, self-driven viewpoint planning and trajectory adaptation.

\subsection{Active Reconstruction and Path Planning Strategies}
While active view planning and Next-Best-View (NBV) offer a significant improvement over a common-used pre-defined flight paths, they introduce new inherent challenges that have become the central to relevant studies \cite{huang2018active, maboudi2023review}. A primary challenge lies in the reliance on a pre-built geometric proxy. Most works still operate in an "\textit{explore-then-exploit}" manner, requiring an initial, often inefficient and onsite flight or existing low-resolution satellite images to generate the coarse model,  upon which subsequent fly path planning depends\cite{zhou2020offsite}. Furthermore, existing NBV planning measures "\textit{reconstructability}" using 
heuristic-based metrics. These handcrafted rules, based on factors like surface coverage or viewing angles, serve as imperfect proxies for final model quality and often fail to balance the goals of real-time performance, reconstruction accuracy and completeness \cite{shen2025aerial}. 

In the last decades, many works have been done to confront these limitations. For example, to replace the "\textit{explore-then-exploit}" manner and improve on-site acquisition efficiency, Zhou et al.\cite{zhou2020offsite} developed an offsite planning method that generated a coarse 2.5D proxy model from 2D maps and satellite imagery, thereby removing the need for an initial onsite exploratory flight. To improve efficiency in scenarios requiring periodic scene updates, Shen et al.\cite{shen2025aerial} proposed a change-aware aerial path planning approach that leverages prior reconstructions and changeability statistics to focus UAV imaging only on regions likely to have changed. Unlike conventional reconstruction-oriented planning methods that uniformly pursue high accuracy and completeness across the entire scene, their strategy selectively revisits only change-prone areas, thereby avoiding unnecessary re-exploration of unchanged regions.

Some Other strategies studies focus on guiding principles of view selection. Focusing on the challenge of sparse views, Ye et al.\cite{ye2024pvp} proposed a progressive system that started with minimal views and iteratively added the most informative one that was determined by a novel metric based on cross-view warping consistency. To better guide the selection process, Gazani et al.\cite{gazani2023bag} explored an appearance-based "\textit{Bag of Views}" approach that selected the next viewpoint by maximizing the expected new visual information for the reconstruction. As a solution for structural integrity, Shang and Shen \cite{shang2023topology} developed a topology-based path planning method that analyzes the global connectivity of a scene to ensure coherent and complete coverage of complex structures.
Nevertheless, to the best of our knowledge, most existing methods are either demonstrated on very small-scale environments or computationally intensive offline planning, leading to insufficient capacity for real-time, in-flight feedback in the case of UAV photogrammetry aiming at surveying large-scale regions. The corresponding gap stems from the absence of a mechanism to leverage an incrementally evolving surface mesh for fine-grained, quality-aware flight path guidance during the on-site flight. 

In this paper, we introduce a mesh-quality aware and predictive system that bridges this gap. It not only evaluates reconstruction quality online but also explores this quality assessment into actionable trajectory updates. By leveraging an incrementally updated mesh and mesh-quality indicators, our work provides fine-grained spatial cues for coverage gaps and directly optimizes UAV flight paths to minimize re-flights and ensure high reconstruction fidelity in large-scale scenes.

\section{Preliminary}

Zhan et al. propose SfM on-the-fly\cite{zhan2025sfm}, achieving near real-time pose estimation and sparse point cloud during image acquisition in arbitrary way. SfM on-the-fly is composed of three modules: first, to deal with newly fly-in unordered images captured in arbitrary way without spatiotemporal continuity, it firstly employs a tailored fine-tuned CNN backbone for extracting global features\cite{hou2023learning} and construct an incremental HNSW (Hierarchical Navigable Small Worlds)\cite{malkov2018efficient} to fast run online image matching; second, to refine the intermediate results and reduce error accumulation, a  hierarchical association tree is built based on the online matching results and a corresponding hierarchical Weighted Local Bundle Adjustment is introduced for online refinement. Finally, to merge multiple sub-models that are generated via various agents, they present an efficient sub-model fusion method using the online image matching for rapidly detecting overlapping information among sub-models and robust 3D similarity transformation to  merge them into a unified framework.  This paper is built on the \textit{SfM on-the-fly} to update poses and sparse point cloud for our subsequent On-the-fly Feedback.

\section{Methodology}

\begin{figure*}[htbp]
    \centering
    \includegraphics[trim={0cm 0 0 0},clip, width=0.95\textwidth]{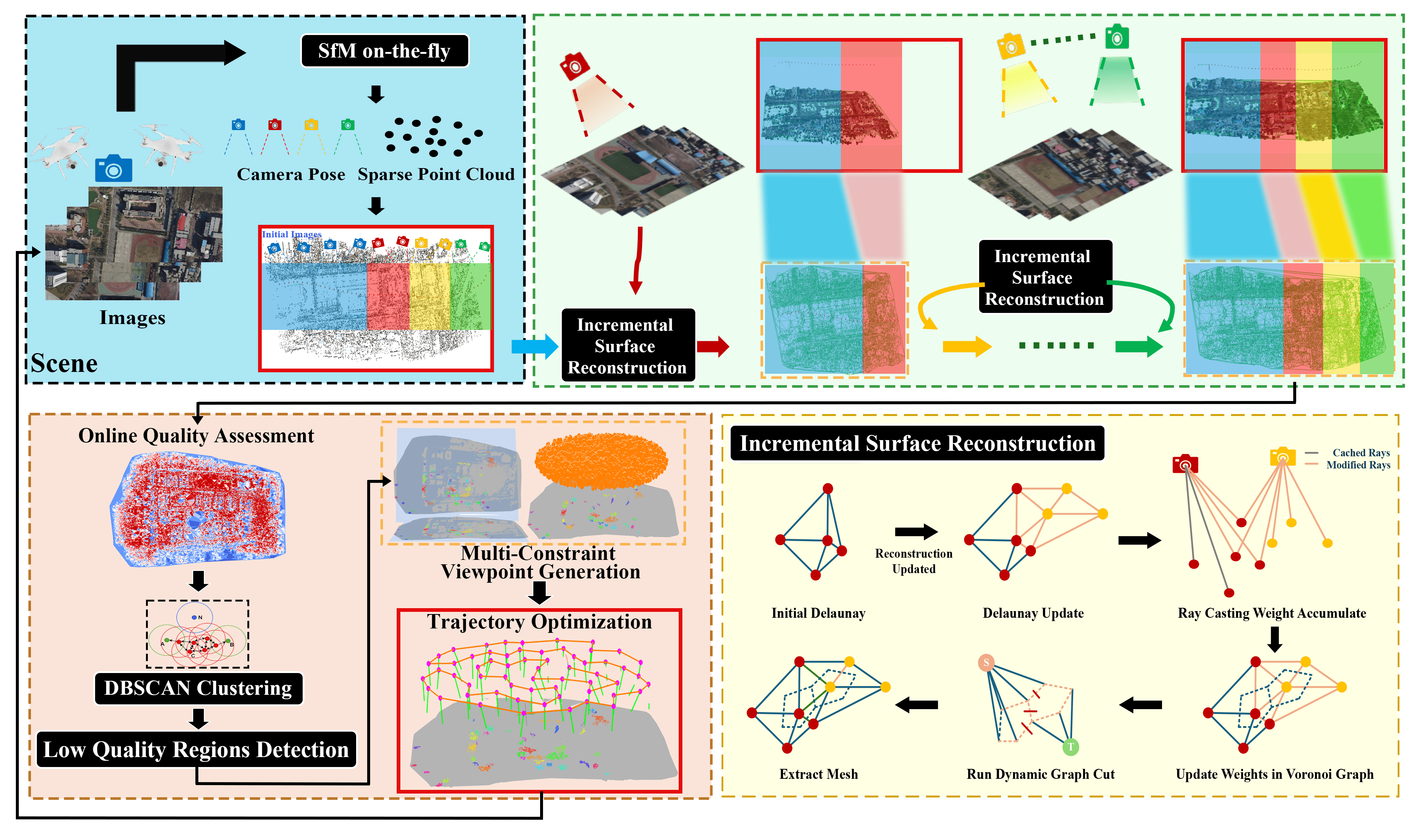}
    \caption{The overall workflow of our On-the-fly Feedback SfM.
Each new batch of UAV images are incrementally processed via \textit{SfM on-the-fly} to refine camera poses and update the sparse point cloud. The evolving sparse cloud is immediately converted into a triangular surface via the incremental surface reconstruction module (yellow panel). This process involves updating the dynamic 3D Delaunay triangulation, accumulating connectivity weights via ray casting, and applying dynamic graph-cut to obtain the optimal partition of tetrahedra into inside and outside space. Online quality assessment computes per-face photogrammetric indicators and identifies low-quality regions, which guide multi-constraint viewpoint generation and real-time trajectory optimization, the resulting trajectory segment is then executed as new fly path by the UAV before the next batch of images arrives.}
    \label{fig:workflow}
     \vspace{-0.6cm}
\end{figure*}

\subsection{Overview of On-the-fly Feedback SfM}
In general, traditional UAV photogrammetry handles all captured images offline, images are first collected via a pre-planned flight path and then processed via a professional software, i.e., "process after capture". Yet, nowadays, real-time UAV photogrammetry that mostly relies on VSLAM can process images during capture, in which image acquisition proceeds also along a pre-planned flight path without awareness of reconstruction quality. This work, in contrast, introduces a fully online and adaptive photogrammetric framework that couples incremental dynamic reconstruction, online reconstruction quality assessment, and predictive path planning into a single continuously operating workflow. The goal is to enable UAVs to improve data acquisition \textit{during} the mission rather than after it.

The overall working pipeline of our method is illustrated in Fig.~\ref{fig:workflow}. It begins with the online pose estimation and sparse point cloud generation using \textit{SfM on-the-fly}, ensuring near real-time SfM results updates without interrupting UAV motion. The evolving point cloud is then converted into a triangular surface mesh via efficient incremental surface reconstruction (see Sect. \ref{incremental suface}), forming a geometry-consistent proxy suitable for early-stage interpretation. Based on the on-going surface mesh, three complementary photogrammetric indicators (including ground sampling distance, observation redundancy, and reprojection error) are computed per triangle face and fused into an ensemble quality score (see Sect. \ref{OQA}), which enables us to automatically expose regions suffering from poor imaging geometry, sparse observation, or high reprojection error. Guided by these online quality cues, our feedback mechanism generates a short-horizon set of candidate viewpoints around the detected low-quality regions through a multi-constraint strategy. These candidate viewpoints are then sparsified under real-time computational constraints and UAV maneuverability considerations, and the resulting subset is organized into a smooth, executable flight segment via lightweight trajectory optimization (See Sect. \ref{PPP_M}). After the UAV executes the planned segment, newly captured images are input into our workflow immediately, which is then used to extend the reconstruction and re-update the flight path again.

Our key idea is to present an integrated \textit{explore–and–exploit} working mode that synchronizes acquisition, evaluation, and planning at short temporal intervals. After the online SfM, instead of operating on each individual image, we process small incoming batches of images (typically 5–20), which provides a balance between computational stability and update responsiveness. In particular, each batch triggers a complete cycle of feedback including incremental surface reconstruction, quality assessment, and trajectory optimization.

To the best of our knowledge, this work is the first fully online photogrammetric framework that unifies image acquisition, incremental surface reconstruction, real-time mesh-quality assessment, and dynamic predictive path planning into a seamless and continuously updated workflow. By tightly synchronizing acquisition and reconstruction, the framework transforms traditional UAV photogrammetry from a passive recording paradigm into an active, quality-aware, and self-planning data collection process.

\subsection{Incremental Surface Reconstruction}\label{incremental suface}
In general, most established methods for reconstructing mesh surfaces from sparse point cloud initiate with a Delaunay triangulation from the input 3D points, after which the resulting tetrahedra are classified as either "inside" or "outside" space via graph-cut optimization on the dual graph to produce a watertight surface along the boundary between these labeled regions \cite{labatut2007efficient}. However, in the context of real-time photogrammetry, the incrementally expanding sparse point cloud from \textit{SfM on-the-fly}\cite{zhan2025sfm} is required to generate a continuous triangular mesh in near real time, and for providing real-time feedback on mesh quality. In general, two primary challenges arise. First, the online SfM incrementally generates new point cloud, meaning that newly added points inevitably affect the regions of the mesh that have been previously constructed. To reconstruct an updated mesh surface, the weights of existing regions are required to be updated, and the dual graph for graph-cut optimization must be rebuilt. This process incurs intensive computational costs that increase sharply with the size of the point cloud. Second, designing an appropriate energy function is crucial for performing dynamic graph cuts to incrementally generate surface mesh. If the update region is too large, redundant computations will grow rapidly; However, if the energy function focuses only on local regions, accumulated weight errors will occur as the scene expands. Based on Labatut et al. \cite{labatut2009robust}, we propose a new energy function to adapt to the growing Delaunay triangulation, taking into account both computational efficiency and the quality of the mesh. Meanwhile, we employ ray casting to track which tetrahedra the rays pass through and mark the changed rays when the point cloud is updated. We also use the dynamic graph-cut method proposed by Kohli et al. \cite{kohli2007dynamic} to update the weights of the energy graph and reuse previous results to compute new graph-cut tasks, thereby greatly improving time efficiency. A simplified workflow can be found at the right bottom of Fig. \ref{fig:workflow}. When a new image is registered, the process begins by dynamically updating the Delaunay tetrahedra with newly acquired 3D points. Subsequently, we compute weights for all newly modified rays, which include both rays connecting existing camera positions to newly added points and rays emanating from the new camera to all its corresponding points. These weights are then accumulated to update the energy function. The updated Delaunay tetrahedra and its associated weights are transformed into a Voronoi graph representation, where vertices correspond to tetrahedron and edges represent shared faces between adjacent tetrahedra. Finally, we apply a dynamic graph cut algorithm on this Voronoi graph to extract the optimal surface mesh.

\subsubsection{Dynamic 3D Delaunay Triangulation}
Newly generated 3D points are incrementally inserted into a dynamically updated 3D Delaunay triangulation. To maintain near real-time performance in large-scale scenes, an efficient incremental insertion algorithm is employed, leveraging spatial locality and hierarchical data structures to minimize computational overhead. For each new point $p$, we first locate and remove the tetrahedron whose circumsphere contains $p$, which forms a cavity. And the resulted cavity is re-triangulated by connecting $p$ to all of its boundary facets.

To maintain Delaunay property, local flip operations are applied to the newly generated tetrahedra until no invalid local structures remain. Efficiency is further enhanced by maintaining an explicit neighbor graph, enabling constant-time traversal between adjacent elements and reducing redundant geometric evaluations. Under the uniform point distributions, the method achieves $O(1)$ insertion time and scales linearly with the size of the affected region, supporting real-time updates in large, dynamically evolving new point cloud.

\begin{figure}[htbp]
    \centering
    \includegraphics[trim={0 0 0 0},clip, width=0.9\linewidth]{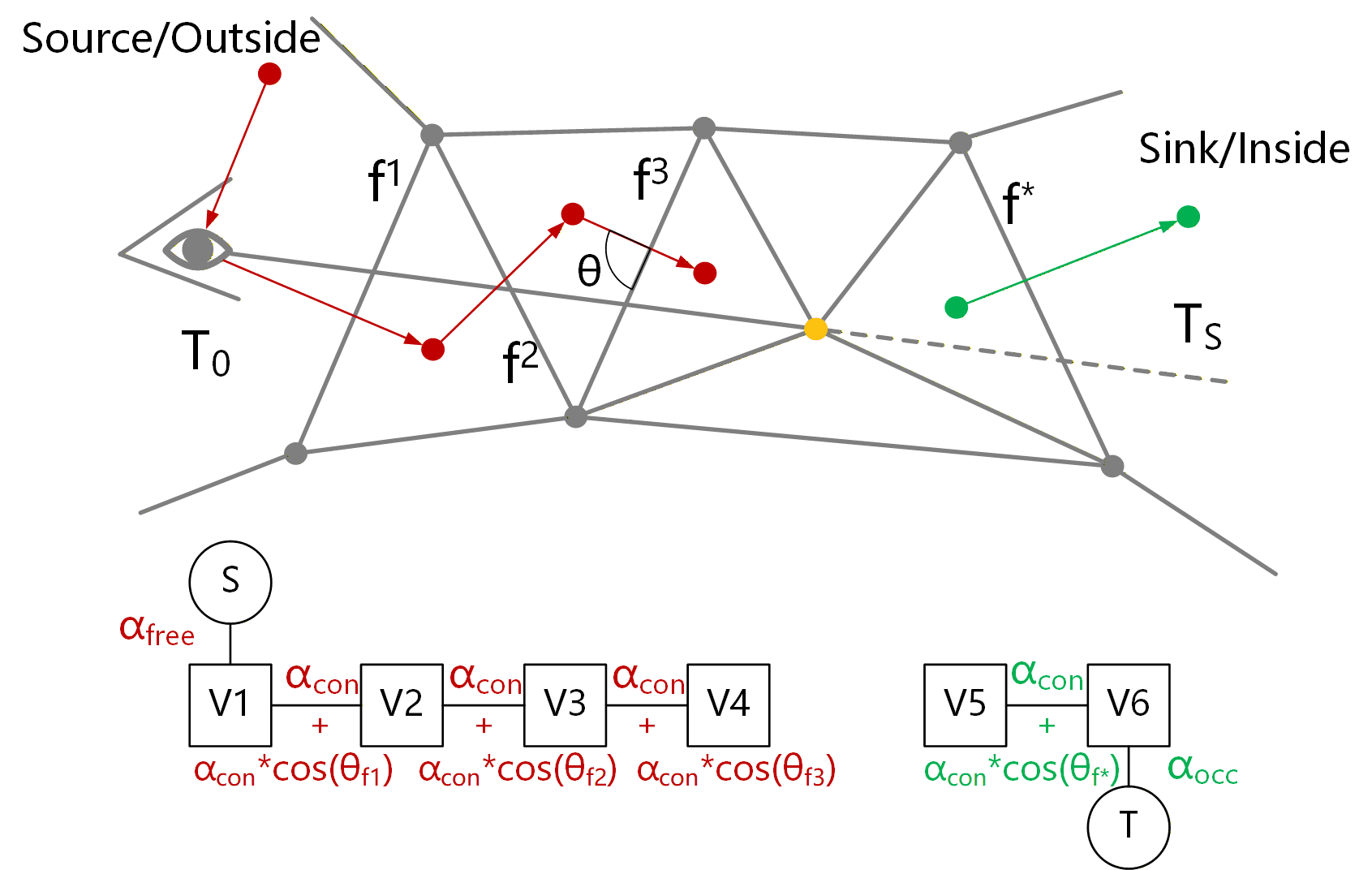}
    \caption{Illustration of the Energy Function model. A line of sight from a camera to an observed point distributes weights to the originating tetrahedron, the intersected facets, and the terminating tetrahedron.}
    \label{fig:soft_visibility}
     \vspace{-0.25cm}
\end{figure}

\subsubsection{Energy Function for Surface Extraction}
Following the triangulation, surface reconstruction is formulated as a binary labeling problem, where each tetrahedron is labeled as inside or outside the surface by minimizing an energy function $E\ (L)$ :

\begin{equation}
E\ (L) = E_{vis}\ (L) + \lambda E_{smooth}\ (L)
\end{equation}

where $E_{vis}$ represents the visibility constraint term, $E_{smooth}$ denotes the regularization of surface geometry, and $\lambda$ is a weight coefficient. $L$ represents the set of rays formed by connecting all camera centers to their corresponding visible 3D points. 

The energy function is mapped into an s-t graph, where each node represents a tetrahedron, and the edges between nodes represent the shared faces between tetrahedra. The two terms of the energy function (1) are interpreted as follows:
\begin{itemize}
\item The visibility term $E_{vis}$ leverages the known camera positions (lines of sight) for each 3D point and it can be formally defined as follows:
\begin{align}
E_{vis}\ (L) = & \alpha_{free} \cdot {L}_{\{T = T_0\}} + \alpha_{con} \cdot \sum_{i=1}^{m} {L}_{\{f = f_i\}} \nonumber \\
 & + \alpha_{con} \cdot {L}_{\{f = f^*\}} 
+ \alpha_{occ} \cdot {L}_{\{T = T_S\}}
\end{align}
For a given 3D point observed by a specific camera, the space along the line of sight between the camera and the point should be outside (\textit{free}), while the space behind the point should be inside (\textit{occupied}). This is encoded into the graph by assigning costs to the nodes (tetrahedra) and edges (triangular facets) that are traversed by these lines of sight. This visibility-driven formulation is highly robust to noise and outliers. As illustrated in Fig.~\ref{fig:soft_visibility}, when a ray incident from the exterior space reaches a 3D point, the first external tetrahedron $T_0$ traversed by the ray should be assigned a connectivity weight $\alpha_{free}$ to the source node. During ray propagation, each triangular facet $f_i$ intersected by the ray path requires corresponding connectivity weights $\alpha_{con}$ to ensure proper weight propagation through the tetrahedral sequence. Meanwhile, the adjacent tetrahedron $T_S$ located behind the 3D point along the ray's extension should be labeled as an internal tetrahedron with weight $\alpha_{occ}$, while their shared face $f^*$ must be assigned the connectivity weight $\alpha_{con}$ to establish complete propagation pathways. This formulation enables continuous visibility modeling in complex occlusion scenarios through explicit characterization of weight assignment mechanisms during ray and tetrahedron interactions. In this paper, the parameters are set as follows: $\alpha_{free}= 1000.0$, $\alpha_{occ} = 1000.0$, $\alpha_{con} = 100.0$.
\item The surface quality term $E_{smooth}$ serves as a regularization term that promotes geometrically plausible surface reconstruction. It is defined as follows:
\begin{align}
E_{smooth} = \alpha_{\text{con}} \cdot \frac{\mathbf{v} \cdot \mathbf{n}_f}{|\mathbf{v}|} = \alpha_{\text{con}} \cdot \cos \theta
\end{align}
considering a shared facet $f$ between adjacent tetrahedra $T_A$ and $T_B$. If the two tetrahedra sharing $f$ have opposite orientation, this indicates that $f$ is likely part of the model's surface region. Based on this geometric characteristic, let $\mathbf{p}_A$ and $\mathbf{p}_B$ denote the vertices of $T_A$ and $T_B$ not incident to $f$. The vector $\mathbf{v} = \mathbf{p}_B - \mathbf{p}_A$ forms an angle $\theta$ with the facet $f$, and facets with small angles are assigned lower weights, making them more likely to be cut during the graph cut optimization. This strategy aims to improve the smoothness of the resulting surface and facilitate the generation of coherent cutting trajectories.
\end{itemize}

When computing each term of the energy function, we take the ray($l$) from the camera center to the 3D point as the fundamental unit. The energy distribution $E(l)$ for each individual ray is first calculated, after which identical energy terms from all rays are aggregated to form the global energy function. The computationally intensive process of accumulating these weights is parallelized via multi-threading to achieve real-time performance.

\textbf{Update Energy Function.}
The energy function $E(L)$ is constructed based on the Delaunay Triangulation ($DT$) structure and visibility information $L$. When the $DT$ topology undergoes modifications or the visibility set $L$ is updated, the energy function requires dynamic recalculation. Our approach treats individual rays as fundamental computational units and introduces a modified rays set to enable incremental computation. This set comprises two categories: newly added visibility rays, and rays affected by structural changes in the $DT$. For each ray in the modified set, we recompute its corresponding energy weight. During the update process, the recomputed weights are merged with historically cached weights through additive accumulation for rays with identical identifiers, thereby achieving efficient energy function updates.

\subsubsection{Dynamic Graph-Cut and Surface Extraction}
After constructing the energy function, we formulate the surface extraction from Delaunay tetrahedralization as a max-flow/min-cut optimization problem. The energy terms are precisely mapped to an s-t graph representation of the Voronoi dual graph, where nodes correspond to tetrahedral cells and edges represent shared triangular faces between adjacent tetrahedra. The minimal cut solution in this graph directly defines the final surface mesh.

To achieve efficient optimization and online incremental surface reconstruction, we employ the dynamic graph cut algorithm proposed by Kohli et al.\cite{kohli2007dynamic}, which maintains a residual flow tree data structure to track incremental changes between successive weight updates. This design ensures that the computational complexity of graph cut optimization depends solely on the number of modified parameters rather than the overall scene scale, thereby enabling real-time surface extraction.

\subsubsection{Mesh Triangle Outlier Filtering}
Based on the introduced method, the resulted mesh often contains irregularly large triangles that compromise geometric fidelity and topological consistency. To enhance mesh quality and mitigate geometric artifacts, an iterative boundary-peeling algorithm integrated with statistical outlier detection is proposed.

In general, in each iteration, boundary triangles (defined as those with at least one edge lacking an adjacent triangle) are identified first. The maximum edge length of all boundary triangles is then collected, and an adaptive threshold is established based on the statistical distribution of these lengths. Specifically, the threshold is defined as $\tau = \mu + k \cdot \sigma$, where $\mu$ represents the mean of the maximum edge lengths, $\sigma$ denotes their standard deviation, and $k$ is an adjustable sensitivity parameter. Per an outlier detection mechanism, triangles whose maximum edge length exceeds the threshold are classified as geometric outliers and removed. This operation exposes previously internal triangles as new boundary elements, thereby triggering subsequent iterations. This iterative process continues until either the maximum number of iterations($n_{it}$) is reached or no outlier of triangular facet can be detected. 


\label{sec:Methodology}
 \begin{figure}[htbp]  
    \centering
    \includegraphics[width=0.99\columnwidth]{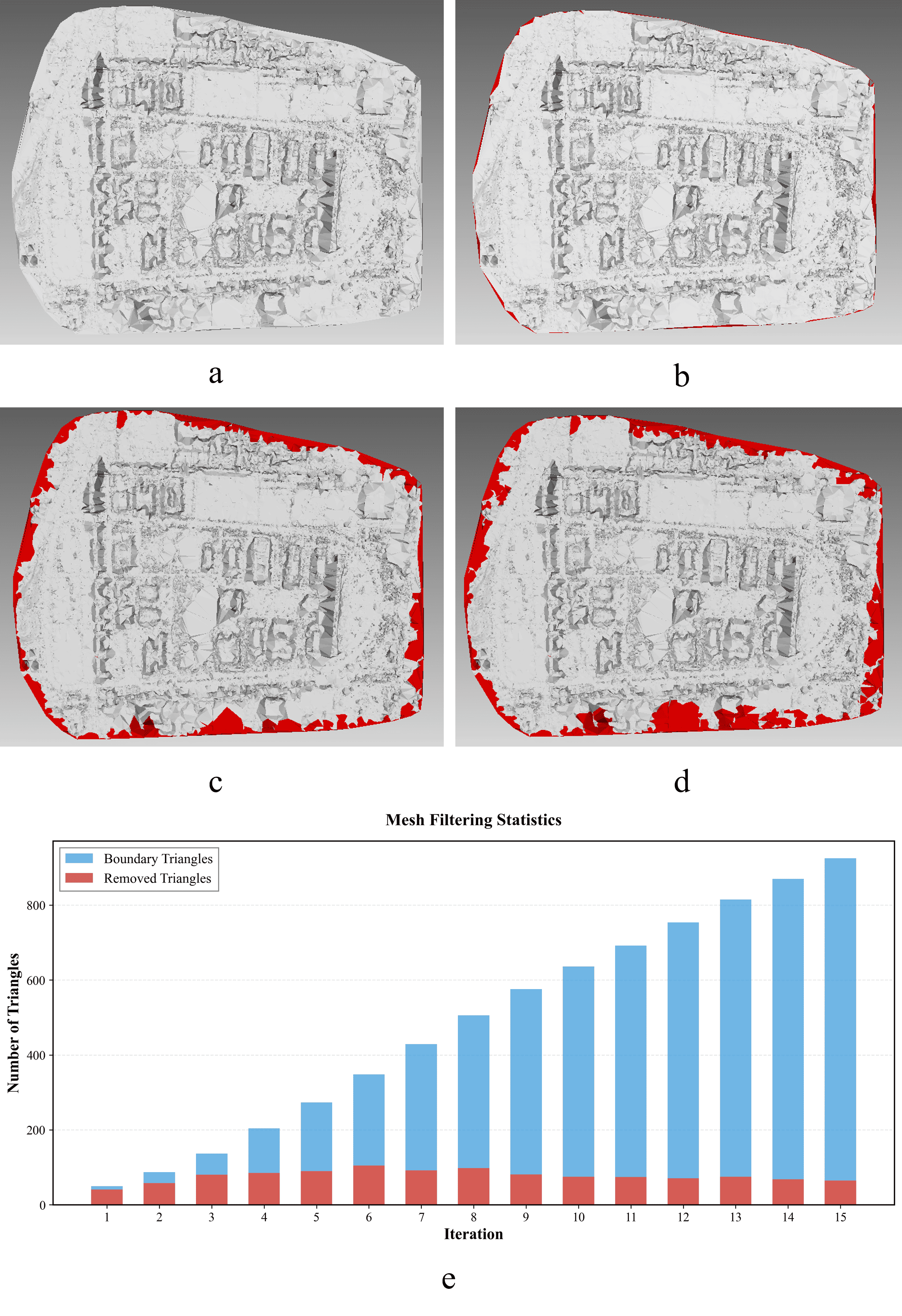}  
    \caption{Comparative results of the Mesh Triangle Outlier Filtering method, (a) Original mesh; (b) $n_{it}=5$; (c) $n_{it}=10$; (d) $n_{it}=15$. Highlighted (red) regions denote the abnormal triangles removed during corresponding iterations. (e) shows mesh filtering statistics in each iteration.}  
    \label{fig:methodology_filter}
    \vspace{-0.3cm}
\end{figure}

Fig. \ref{fig:methodology_filter} illustrates the performance of mesh triangle outlier filtering on the \textit{PHANTOM} dataset (see Sect. \ref{sec:experiment_datasets}). From Fig. \ref{fig:methodology_filter}(a) to Fig. \ref{fig:methodology_filter}(d), the input mesh and refined meshes after 5, 10, and 15 iterations of filtering are shown. At $n_{it}=5$, our method can successfully eliminate most irregularly elongated triangles while producing smooth boundaries that preserve structural integrity. 
When $n_{it}$ increases to 10 and 15, while extra outlier removal is achieved, it introduces over-smoothing artifacts—manifested as isolated mesh patches and aliased boundaries. This phenomenon significantly compromises both the structural integrity of the topology and the precision of geometric representation, which can be explained by the fact that excessive iterations disrupt the natural curvature distribution along mesh boundaries, leading to over trimming of the characteristic edges that should be preserved.

Fig. \ref{fig:methodology_filter}(e) illustrates the quantitative evolution of filtered triangles and boundary triangles during 15 consecutive iterations. It demonstrates that the number of removed triangles initially increases, peaks at the 6th iteration (105 triangles), and subsequently declines. Meanwhile, the count of boundary triangles exhibits monotonic growth, reflecting the algorithm's mechanism of progressively exposing internal triangles. Notably, the decline in removal count after the 5th iteration suggests that primary outlier regions have been addressed, with diminishing returns in subsequent steps. Integrating qualitative observations, we identify $n_{it}=5$ and $k=2.0$ (the default setting in this  work) as the optimal parameters, achieving the best trade-off between efficiency and reconstruction quality.



\subsection{Online Quality Assessment} \label{OQA}
The evolving mesh not only provides an interpretable geometric proxy for an initial and coarse assessment of the on-the-go 3D model, but also serves as the basis for continuous online quality evaluation, where three complementary indicators are estimated for each triangle mesh to reveal the reconstruction quality. In this work, three criteria are considered: 

\textbf{(1) Ground Sampling Distance (GSD).}
The GSD measures the effective spatial resolution of a reconstructed surface element with respect to the captured images\cite{hoppe2012online}. For each triangular face $f_i$ in the mesh, we reproject it onto every corresponding aligned view $C_t$, and determine the projected 2D image area $P(f_i, C_t)$ (in pixels) and the 3D area $A(f_i)$. The GSD for each triangular face is then defined as:
\begin{equation}
\mathrm{GSD}_i = \min_{t \in \mathcal{C}_{vis}(f_i)} 
\sqrt{\frac{A(f_i)}{P(f_i, C_t)}}
\end{equation}
where $\mathcal{C}_{vis}(f_i)$ denotes the set of views on which the triangular face $f_i$ is visible (i.e., not self-occluded). The lower the $\mathrm{GSD}_i$ is, the higher the effective ground resolution should be, indicating sharper detail and denser image sampling. In practice, we leverage GPU-based rasterization to efficiently estimate $P(f_i, C_t)$ with an accurate visibility check.

\textbf{(2) Observation Redundancy.}
For a given surface triangle, this metric reflects the number of views that can be observed, which directly correlates with the robustness of the 3D reconstruction quality. For triangle $f_i$, the observation redundancy is defined as:
\begin{equation}
R_i = |\mathcal{C}_{vis}(f_i)|
\end{equation}
where $\mathcal{C}_{vis}(f_i)$ is the same visible view set as above. A higher redundancy implies stronger multi-view geometric constraints, thereby enhancing triangulation stability.

\textbf{(3) Reprojection Error.}
The reprojection error quantifies the geometric consistency between the reconstructed surface and the image observations. For each vertex $v_j$ of the triangular face $f_i$, its corresponding 3D position is projected back into each visible image $C_t$, and the reprojection error $\mathbf{p}_{t,j}$ is measured. The per-triangle reprojection error is then computed as the mean over all vertices and visible images:
\begin{equation}
E_i = 
\frac{1}{3 \, |\mathcal{C}_{vis}(f_i)|} 
\sum_{v_j \in f_i} 
\sum_{C_t \in \mathcal{C}_{vis}(f_i)}
\left\| \pi_{C_t}(v_j) - \mathbf{p}_{t,j} \right\|_2
\end{equation}
where $\pi_{C_t}(\cdot)$ denotes the projection function of camera $C_t$. 
Lower $E_i$ values indicate better geometric alignment between reconstructed surfaces and image observations.

\textbf{(4) Ensemble Mesh Quality Indicator.} 
For each triangular face $f_i$ in the evolving mesh $\mathcal{M}$, the three individual indicators are first normalized and then fused into a single ensemble quality index, denoted as $Q_{total}^i$. 
\begin{equation}
\small
Q_{total}^i = w_{GSD} \cdot \mathcal{N}(GSD_i^{-1})   
+ w_R \cdot \mathcal{N}(R_i) 
+ w_E \cdot \mathcal{N}(E_i^{-1})
\end{equation}

where $\mathcal{N}(\cdot)$ denotes percentile-based normalization to the range $[0,1]$. In our implementation, we empirically set the weights as follows: ${w_{GSD}}=0.1$, $w_R=0.8$, $w_E=0.1$. The inverse transformations ensure that higher $Q_{total}$ values correspond to better reconstruction quality.

\begin{figure*}[htbp]
    \centering
    \includegraphics[trim={0cm 0 0 0},clip, width=0.95\textwidth]{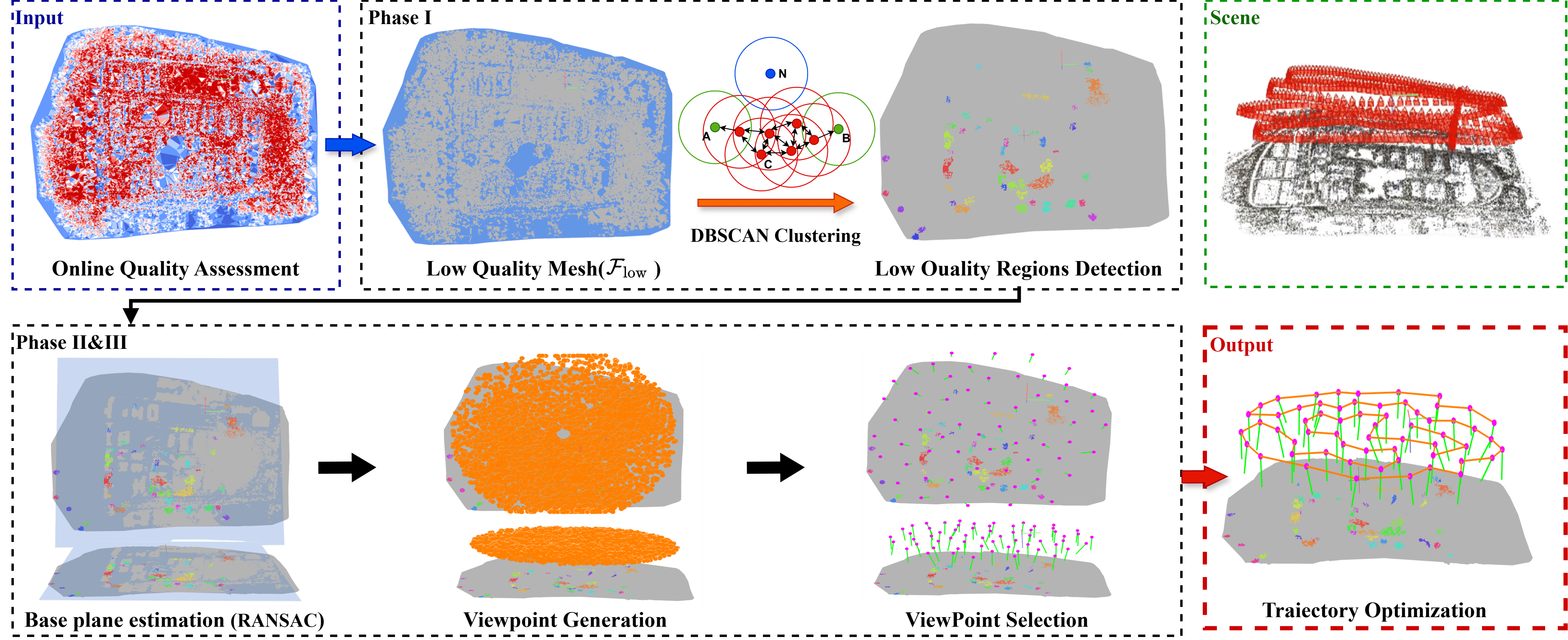}
    \caption{
Workflow of the proposed predictive path planning pipeline. It contains five steps: 
(a) Low-quality regions are identified based on the per-face ensemble quality score $Q_{total}$. 
(b) DBSCAN clustering groups spatially adjacent low-quality faces into coherent regions, with distinct colors indicating different clusters. 
(c) Candidate viewpoints are generated around each region under multi-constraint sampling. 
(d) A sparsification step selects a compact, well-distributed subset of viewpoints. 
(e) The final trajectory is optimized for smooth and efficient UAV flight.
}
    \label{fig:path_palnning}
     \vspace{-0.25cm}
\end{figure*}

\subsection{Real-Time Predictive Path Planning with Mesh-quality Aware Trajectory Optimization} \label{PPP_M}
To empower the UAV to actively explore the environment and optimize the image acquisition in real-time, a predictive path planning module that continuously adapts the UAV trajectory based on mesh-quality indicators is designed. As Fig. \ref{fig:path_palnning} illustrates, this module consists of four parts: (i) low-quality regions detection, (ii) multi-constraint viewpoint generation, (iii) viewpoint selection, and (iv) trajectory optimization.

\subsubsection{Low-Quality Regions Identification}
The path planning process begins by identifying regions on the evolving mesh $\mathcal{M}$ that exhibit low reconstruction quality and thus require additional consideration. This is achieved by leveraging the ensemble quality index $Q_{total}^i$, which is computed for each triangular face $f_i$ as detailed in the preceding section \ref{OQA}.

Based on a self-adaptive threshold $\tau_{quality}$, faces with insufficient reconstruction quality are then filtered into a set, defined as $\mathcal{F}_{\text{low}}$, which effectively targets the poorly reconstructed areas for subsequent refinement:
\begin{equation}
\mathcal{F}_{\text{low}} = \{ f_i \in \mathcal{M} \mid Q_{total}^i \leq \tau_{quality} \}
\end{equation}

The filtered mesh faces in $\mathcal{F}_{\text{low}}$ correspond to areas with possible insufficient image redundancy, less view observation, or high reprojection error that collectively reduce local reconstruction reliability. To transform these discrete faces into coherent regions, we apply DBSCAN clustering \cite{ester1996density} on their corresponding centroids $\{\mathbf{c}_i\}$, grouping spatially adjacent low-quality faces into a set of clusters $\mathcal{G}$:
\begin{equation}
\mathcal{G} = \text{DBSCAN}\bigl(\{\mathbf{c}_i : f_i \in \mathcal{F}_{\text{low}}\}, \epsilon_{\text{spatial}}, N_{min}\bigr),
\end{equation}
where $\epsilon_{\text{spatial}} = 0.008 \times D_{\text{scene}}$ ($D_{\text{scene}}$ is the scene diagonal) is the neighborhood radius and $N_\text{min}$ is the adaptive minimum cluster size. Each resulting cluster $\mathcal{G}_k$ represents a spatially continuous low-quality region that requires additional planning for the image acquisition.

\subsubsection{Multi-Constraint Viewpoint Generation}
With the identified low-quality regions, we first generate a large pool of candidate viewpoints to ensure comprehensive coverage and quality improvement. This process begins by establishing a consistent spatial reference via a robust plane fitting (based on RANSAC\cite{fischler1981random}) on the mesh vertices, defining a base plane $(\mathbf{n}_{\text{plane}}, \mathbf{p}_{\text{plane}})$:
\begin{equation}
 (\mathbf{n}_{\text{plane}}, \mathbf{p}_{\text{plane}}) = \text{RANSAC}_{\text{plane}}(\{\mathbf{v}_i\}, \epsilon_{\text{dist}}, N_{\text{iter}})
\end{equation}
where $\mathbf{n}_{\text{plane}}$ is the plane normal, $\mathbf{p}_{\text{plane}}$ is a point on the plane, $\epsilon_{\text{dist}}$ is the distance threshold, and $N_{\text{iter}}=100$ is the maximum iterations. If mesh-based fitting fails, we alternatively fall back to camera center fitting, and ultimately to a default horizontal plane using the scene centroid.

For each low-quality cluster $\mathcal{G}_k$, its spatial centroid $\mathbf{g}_k$ as the mean position of all face centroids within the cluster is first computed:
\begin{equation}
\mathbf{g}_k = \frac{1}{|\mathcal{G}_k|} \sum_{f_i \in \mathcal{G}_k} \mathbf{c}_i.
\end{equation} 
where $\mathbf{c}_i$ is the centroid of face $f_i$.

Instead of simple uniform angular sampling, we adopt a \textit{quality-weighted normal sampling strategy}
that leverages the inherent directional information encoded in the surface normals of low-quality faces. 
Each face $f_i \in \mathcal{G}_k$ contributes a candidate viewing direction $\mathbf{n}_i$ 
(the face normal), weighted by its reconstruction quality deficit: $w_i = \max(0, \tau_{\text{quality}} - Q_{\text{total}}^i)$,
where $Q_{\text{total}}^i$ is the quality score of face $f_i$ and $\tau_{\text{quality}}$ 
is the quality threshold. Faces with lower quality contribute higher weights, 
prioritizing viewing directions toward the most deficient regions.

For each face $f_i$ in the cluster, a ray is cast from the cluster centroid $\mathbf{g}_k$ 
along the face normal direction:
\begin{equation}
\mathbf{r}_j(t) = \mathbf{g}_k + t \cdot \mathbf{n}_i, \quad t \geq 0.
\end{equation}

To ensure that the final viewpoints lie within a valid and UAV-feasible 3D spatial domain,
an oriented bounding box (OBB), aligned with the fitted base plane 
$\mathbf{n}_{\text{plane}}$ and incrementally enclosing the scene, is constructed. 
The construction begins by computing the axis-aligned bounding box (AABB) center of all mesh vertices: $\mathbf{b}_{\text{center}}$ = $\tfrac{1}{2}$$(\mathbf{p}_{\min}+\mathbf{p}_{\max})$,
where $\mathbf{p}_{\min}$ and $\mathbf{p}_{\max}$ denote the minimum and maximum vertex coordinates, respectively.

This center is then projected onto the base plane to obtain the OBB origin:
\begin{equation}
d = (\mathbf{b}_{\text{center}} - \mathbf{p}_{\text{plane}})\cdot\mathbf{n}_{\text{plane}}, 
\mathbf{b}_{\text{origin}} = \mathbf{b}_{\text{center}} - d\,\mathbf{n}_{\text{plane}}.
\end{equation}

A local orthonormal coordinate frame $(\mathbf{u},\mathbf{v},\mathbf{n})$ is then defined,
where $\mathbf{n}=\mathbf{n}_{\text{plane}}$ and
\begin{equation}
\mathbf{u} = 
\frac{\mathbf{a}\times\mathbf{n}}{\|\mathbf{a}\times\mathbf{n}\|},\qquad
\mathbf{v} = \mathbf{n}\times\mathbf{u}.
\end{equation}

In this coordinate frame, each mesh vertex $\mathbf{p}_i$ is projected as:
\begin{equation}
\begin{aligned}
u_i = (\mathbf{p}_i - \mathbf{b}_{\text{origin}})\cdot\mathbf{u}, \\
v_i = (\mathbf{p}_i - \mathbf{b}_{\text{origin}})\cdot\mathbf{v}, \\
h_i = (\mathbf{p}_i - \mathbf{b}_{\text{origin}})\cdot\mathbf{n}.
\end{aligned}
\end{equation}

The box extents are then computed by taking the min/max values along each axis:
\begin{equation}
\begin{aligned}
u_{\min} &= \min_i u_i,\;\; u_{\max} = \max_i u_i, \\
v_{\min} &= \min_i v_i,\;\; v_{\max} = \max_i v_i, \\
h_{\min} &= \min_i h_i,\;\; h_{\max} = \max_i h_i.
\end{aligned}
\end{equation}

Using a scaling factor $\alpha$ (typically $0.8$), the OBB center in local coordinates is given by
\begin{equation}
\begin{aligned}
c_u &= \tfrac{1}{2}(u_{\min}+u_{\max}), \quad \text{extent}_u = (u_{\max}-u_{\min})\alpha,\\
c_v &= \tfrac{1}{2}(v_{\min}+v_{\max}), \quad \text{extent}_v = (v_{\max}-v_{\min})\alpha,\\
c_h &= \tfrac{1}{2}(h_{\min}+h_{\max}), \quad \text{extent}_h = (h_{\max}-h_{\min})\alpha.
\end{aligned}
\end{equation}

Finally, the OBB bounds in the local coordinate frame are expressed as:
\begin{align}
\mathcal{B}_{\text{OBB}} = \{\,
&c_u \in [c_u - \tfrac{\text{extent}_u}{2},\;\; c_u + \tfrac{\text{extent}_u}{2}], \nonumber\\
&c_v \in [c_v - \tfrac{\text{extent}_v}{2},\;\; c_v + \tfrac{\text{extent}_v}{2}], \\
&c_h \in [c_h - \tfrac{\text{extent}_h}{2},\;\; c_h + \tfrac{\text{extent}_h}{2}]
\,\}. \nonumber
\end{align}

With the defined OBB, each sampled ray is intersected with the bounding box to obtain the
actual candidate viewpoints. The ray is first transformed into the OBB coordinate frame:
\begin{equation}
\mathbf{r}_{\text{local}}(t)=
\begin{bmatrix}
(\mathbf{r}_j(t)-\mathbf{b}_{\text{origin}})\cdot\mathbf{u} \\
(\mathbf{r}_j(t)-\mathbf{b}_{\text{origin}})\cdot\mathbf{v} \\
(\mathbf{r}_j(t)-\mathbf{b}_{\text{origin}})\cdot\mathbf{n}
\end{bmatrix},
\end{equation}
and the intersection parameter $t_{\text{hit}}$ is solved such that
$\mathbf{r}_{\text{local}}(t_{\text{hit}}) \in \mathcal{B}_{\text{OBB}}$.
If a valid intersection exists, the viewpoint position is:
\begin{equation}
\mathbf{v}_j^{\text{candidate}}
= \mathbf{r}_j(t_{\text{hit}})
= \mathbf{g}_k + t_{\text{hit}} \cdot \mathbf{d}_j .
\end{equation}

This approach ensures that all candidate viewpoints remain within a compact, terrain-aligned, and UAV-feasible envelope. Combined with \textbf{\textit{multi-scale sampling}} strategy, it can yield a diverse and well-distributed set of candidate viewpoints for subsequent viewpoint selection and trajectory optimization.

\subsubsection{Viewpoint Selection} 
From the large pool of valid candidates, the greedy sparsification algorithm is employed to select the compact and final set of viewpoints $\mathcal{V}_{\text{final}}$ while ensuring real-time performance. 
To achieve both spatial diversity and computational efficiency, the selection process is designed with bounded complexity and adaptive control over sampling density.

The candidate viewpoint is iteratively selected, which maximizes the minimum distance to all previously selected viewpoints, ensuring a well-distributed spatial configuration:
\begin{equation}
 \mathbf{v}_{\text{next}} = \arg\max_{\mathbf{v} \in \mathcal{V}_{\text{remaining}}} 
 \min_{\mathbf{u} \in \mathcal{V}_{\text{selected}}} \|\mathbf{v} - \mathbf{u}\|,
\end{equation}
subject to a minimum spacing constraint $\|\mathbf{v}_{\text{next}} - \mathbf{u}\| \geq d_{\text{min}}, \forall \mathbf{u} \in \mathcal{V}_{\text{selected}}$.
This process effectively reduces viewpoint redundancy while preserving the overall reconstruction performance.

To maintain online responsiveness under large candidate sets, the sparsification process integrates several mechanisms designed to bound computation while preserving spatial representativeness:

\textbf{Adaptive Target Size.} 
The number of selected viewpoints is adaptively adjusted according to the candidate pool size, ensuring that the final viewpoint set remains compact and computationally tractable. 
This adaptive adjustment keeps the runtime roughly constant across varying scene complexities.

\textbf{Scene-Adaptive Minimum Spacing.}
A dynamic minimum distance threshold, determined by the overall scene scale, 
prevents oversampling and ensures sufficient spatial diversity among selected viewpoints. 
Once no remaining candidates satisfy the spacing requirement, the selection process terminates automatically.

\textbf{Bounded Greedy Iteration.} 
The greedy selection inherently converges early in practice, as the adaptive spacing constraint quickly filters out redundant candidates in densely sampled regions, 
further constraining runtime without compromising viewpoint distribution quality.



\subsubsection{Trajectory Optimization}
The final stage sequences the selected viewpoints $\mathcal{V}_{\text{final}}$ into an efficient, flyable path. An improved nearest-neighbor method generates an initial trajectory, guided by a UAV-specific cost function that penalizes altitude changes to ensure a smooth flight:
\begin{equation}
 C(\mathbf{v}_i, \mathbf{v}_j) = |\mathbf{v}_i - \mathbf{v}_j| + 0.3 \cdot |z_j - z_i|
\end{equation}
This initial path is then further refined using 2-opt local optimization \cite{croes1958method} to minimize the total flight distance while maintaining operational efficiency, yielding the final adaptive trajectory for the UAV.

\section{Experiments}
In this section, comprehensive experiments are reported to evaluate the proposed online feedback photogrammetric framework in terms of reconstruction quality, online responsiveness, and adaptive path planning performance. The objectives of the conducted experiments are fourfold: (i) exploring the performance of our incremental mesh generation; (ii) verifying the reliability of the proposed online mesh-quality indicator in capturing the evolving reconstruction state during incremental image acquisition; (ii) investigating our predictive, quality-driven path planning strategy that can effectively reallocate viewpoints toward low-quality regions and improve scene coverage; (iv) most importantly, demonstrating the overall efficiency and practicality of the our "explore–and-exploit" feedback pipeline. 

To this end, both self-captured and public UAV datasets are employed (see Sect.~\ref{sec:experiment_datasets}), and evaluations are conducted across four complementary perspectives: surface reconstruction, online quality assessment, adaptive path planning and the entire "explore-and-exploit" feedback pipeline. All experiments are run on a machine with i9-12900 K CPU and RTX3080 GPU.

\subsection{Experimental Setup}

\begin{table*}[htbp]
\centering
\caption{Details of experimental datasets.}
\label{tab:dataset}
\vspace{-0.2cm}

\renewcommand{\arraystretch}{1.2}

\begin{tabularx}{\textwidth}{ 
    >{\centering\arraybackslash}X 
    >{\centering\arraybackslash}X 
    >{\raggedright\arraybackslash}X
    >{\centering\arraybackslash}X 
    >{\centering\arraybackslash}X
}
\toprule
Dataset & Image Number & Source & Capture Platform & Resolution \\
\midrule
\textit{SHHY}     & 770 & Self-Captured & \textit{DJI Mavic2 pro} & 1920 $\times$ 1080 \\
\textit{PHANTOM}  & 467 & Bu et al., 2016\cite{7759672} & \textit{DJI Mavic2 pro} & 1920 $\times$ 1080 \\
\textit{US3D}     & 990 & Lin et al., 2022\cite{UrbanScene3D} & Public dataset & 5472 $\times$ 3648 \\
\textit{GYM}  & 580 & Self-Captured & \textit{DJI Matrice 4T} & 4032 $\times$ 3024 \\
\textit{YS}  & 320 & Self-Captured & \textit{DJI Matrice 4T} & 4032 $\times$ 3024 \\
\textit{XingHu}  & / & Self-Captured & \textit{DJI Matrice 4T} & 4032 $\times$ 3024 \\
\bottomrule
\end{tabularx}

\vspace{-0.3cm}
\end{table*}

\label{sec:Experiments}
 \begin{figure*}[!t]  
    \centering
    \includegraphics[width=0.95\textwidth]{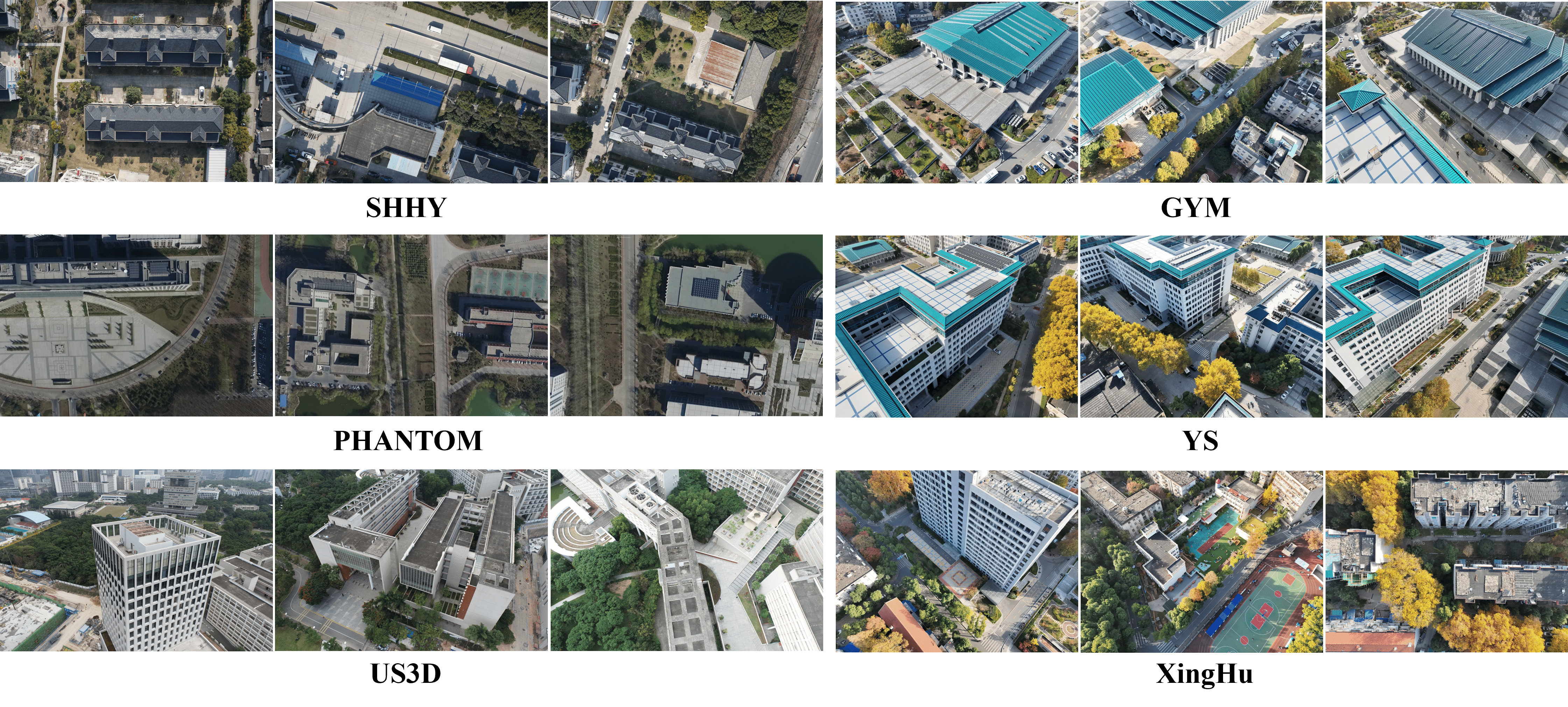}  
    \caption{Sample images of experimental datasets.}  
    \label{fig:data}
\end{figure*}

\subsubsection{Experimental Datasets}\label{sec:experiment_datasets}
Several UAV datasets (see sample images in Fig.~\ref{fig:data}) from various real-world scenarios are tested to evaluate the proposed online feedback photogrammetric framework. Four datasets (\textit{SHHY, GYM, YS, XingHu}) are self-captured by the UAV platform of  \textit{DJI Mavic 2 Pro} and \textit{DJI Matrice 4T}. 
The \textit{PHANTOM} dataset, captured using \textit{DJI Mavic 2 Pro}, is from Du et al.~\cite{11126873}, which features high forward overlap (approximately 92\%).
The dataset (\textit{US3D}) is from a publicly available UrbanScene3D benchmark \cite{UrbanScene3D}, which consists of a large-scale aerial–ground urban reconstruction dataset frequently used in path planning, scene understanding, and active reconstruction studies. UrbanScene3D provides high-resolution imagery, LiDAR scans, and fine-grained urban geometry captured across diverse metropolitan environments, offering complex structural variations suitable for evaluating viewpoint selection and trajectory optimization strategies. 
(More details can be found at Tab.~\ref{tab:dataset})

In particular, \textit{SHHY, GYM} and \textit{YS} are employed to test our proposed incremental surface reconstruction method in Sect.~\ref{sec:mesh_exp}, which contains evaluations of mesh quality and time efficiency, \textit{PHANTOM} and \textit{US3D} are used to validate the performance of our proposed online mesh-quality indicator in Sect.~\ref{sec:quality_assessment_exp}, \textit{US3D} is applied to investigate our mesh-quality-driven path planning strategy in Sect.~\ref{sec:path_planning_exp}, the self-captured \textit{XingHu} is used to demonstrate the overall efficiency and practicality of our "explore–and-exploit" feedback pipeline in Sect.~\ref{sec:entire_pipeline_exp}.

\subsubsection{Evaluation Metrics}
\label{sec:exp_metrics}



To quantitatively evaluate the quality of the generated mesh, we adopt the widely used precision–recall formulation from Knapitsch et al. \cite{knapitsch2017tanks}. 
Let $\mathcal{G}$ denote the ground-truth point set and $\mathcal{R}$ denote the reconstructed point set extracted from our mesh. 

For each reconstructed point $r \in \mathcal{R}$, its distance to the ground truth is defined as
\begin{equation}
e_{r \rightarrow \mathcal{G}} = \min_{g \in \mathcal{G}} \| r - g \| .
\end{equation}

Given a distance threshold $d$, the \textit{precision} $P(d)$ measures the percentage of reconstructed points that lie within distance $d$ from the ground truth:
\begin{equation}
P(d) = 
\frac{100}{|\mathcal{R}|}
\sum_{r \in \mathcal{R}} 
\left[\, e_{r \rightarrow \mathcal{G}} < d \,\right],
\end{equation}
where $[\,\cdot\,]$ denotes the Iverson bracket, which evaluates to $1$ if the condition is true and $0$ otherwise.

Similarly, for each ground-truth point $g \in \mathcal{G}$, its distance to the reconstruction is defined as
\begin{equation}
e_{g \rightarrow \mathcal{R}} = \min_{r \in \mathcal{R}} \| g - r \| .
\end{equation}

The \textit{recall} $R(d)$ is defined as the percentage of ground-truth points that can be explained by the reconstruction within distance $d$:
\begin{equation}
R(d) = 
\frac{100}{|\mathcal{G}|}
\sum_{g \in \mathcal{G}}
\left[\, e_{g \rightarrow \mathcal{R}} < d \,\right].
\end{equation}

Finally, precision and recall are combined into a single summary measure using the F-score:
\begin{equation}
F(d) = 
\frac{2 P(d) R(d)}{P(d) + R(d)}.
\end{equation}

A higher F-score can reflect the mesh model that is simultaneously accurate (high precision) and complete (high recall). 
Following the standard Tanks and Temples evaluation protocol, we report $P(d)$, $R(d)$, and $F(d)$ at a fixed threshold $d$ for all relevant experiments.


\subsubsection{Time Efficiency}
Time efficiency is very crucial for online processing, which is evaluated in terms of the \textit{average processing time per image}, providing a direct measure of computational cost during dynamic incremental reconstruction and online feedback.  

\subsection{Performance of Our Incremental Surface Reconstruction}\label{sec:mesh_exp}

In this section, we conduct two tests. First, to evaluate the mesh quality, taking the mesh model generated by the commercial software - ContextCapture (CC) as reference, our proposed incremental coarse surface reconstruction is compared with two mainstream SfM pipelines (Colmap \cite{schonberger2016structure} and openMVG \cite{moulon2016openmvg}), and their sparse point clouds are used to generate coarse mesh for a fair comparison. Second, in terms of time efficiency, we test the feasibility of the proposed energy function and online incremental processing for dynamically extracting surfaces from real-time extended point clouds, whose complexity is typically independent of the overall scene scale. And, a relevant work of Labatut et al.\cite{labatut2009robust} is compared.
\subsubsection{Evaluation of mesh quality}
Based on datasets of \textit{SHHY}, \textit{GYM}, \textit{YS}, both qualitative and quantitative evaluations of generated coarse mesh model are investigated by comparing with COLMAP, OpenMVG. First, the spatial resolution of the reference point clouds—generated using CC for each dataset—is normalized through down-sampling. Specifically, voxel grid filtering is applied to the reference point clouds to match the point density of the corresponding reconstructed point clouds. For quantitative assessment, we follow the evaluation protocol established in~\cite{knapitsch2017tanks}, computing  Accuracy($P(d)$), Completeness($R(d)$) and F1 score($F(d)$), as defined in Section~\ref{sec:exp_metrics}.

\begin{table}[htbp]
\centering
\caption{Quantitative comparison of three surface reconstruction methods on three tested datasets.Bold indicates the best performance, underline indicates the second-best metrics}
\label{tab:quantitative_mesh_quality}
\vspace{-0.2cm}
\renewcommand{\arraystretch}{1.2}

\begin{tabular}{ccccc}
\hline
Dataset & Method & Accuracy & Completeness & F1 Score \\
\hline
             & Colmap & 0.4970 & \textbf{0.4771} & 0.4868 \\
\textit{SHHY}& OpenMVG & \textbf{0.7270} & 0.3633 & 0.4845 \\
             & Ours & \uuline{0.6509} & \uuline{0.4527} & \textbf{0.5340} \\
\hline
             & Colmap & \uuline{0.7793} & \textbf{0.6092} & \textbf{0.6838} \\
\textit{GYM}& OpenMVG & 0.7363 & 0.4937 & 0.5911 \\
             & Ours & \textbf{0.8001} & \uuline{0.5958} & \uuline{0.6830} \\
\hline
             & Colmap & \uuline{0.7042} & \textbf{0.5755} & \textbf{0.6334} \\
\textit{YS}& OpenMVG & 0.7054 & 0.4672 & 0.5621 \\
             & Ours & \textbf{0.7271} & \uuline{0.5533} & \uuline{0.6284} \\
\hline
\end{tabular}
\vspace{-0.3cm}
\end{table}

\begin{figure}[htbp]  
    \centering
    \includegraphics[width=0.9999\linewidth]{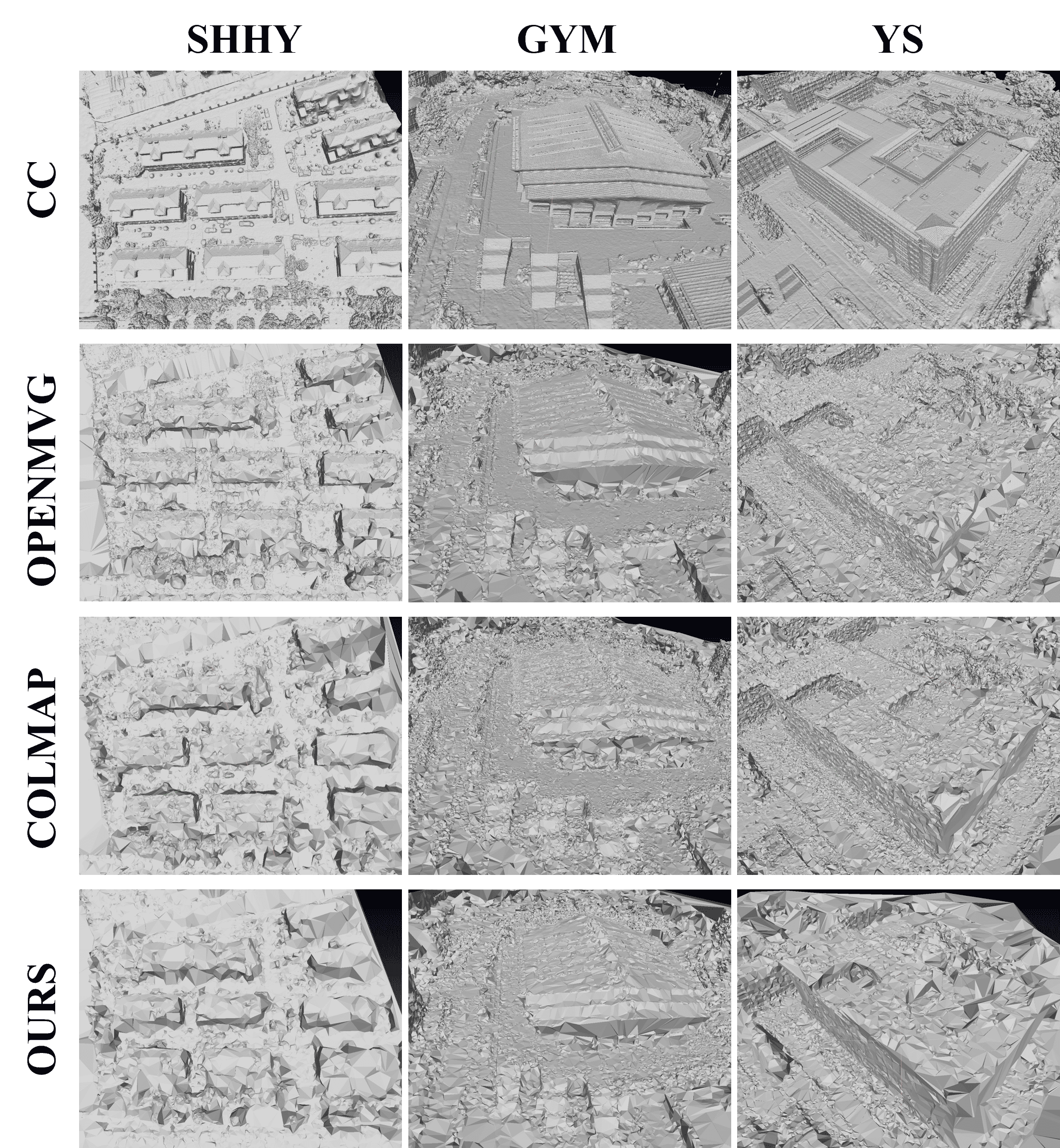}
    \caption{Comparison of mesh model generated by different methods. The results of CC (ContextCapture) is used as reference.}  
    \label{fig:mesh_gym_roiqualitycompare}
    \vspace{-0.3cm}
\end{figure}

From Tab. ~\ref{tab:quantitative_mesh_quality}, it is evident that the proposed incremental surface reconstruction method can achieve competitive performance across all datasets, consistently yielding either the best or second-best results in most metrics, despite not universally outperforming all baselines in every category. On the \textit{SHHY}, our approach can attain the highest F1 score (0.5340), indicating better overall balance between accuracy and completeness. While OpenMVG yields higher accuracy (0.7270), it suffers from notably lower completeness (0.3633), revealing a less complete reconstruction. Similarly, on the \textit{GYM} and \textit{YS} datasets, our method achieves leading accuracy (0.8001 and 0.7271, respectively), while maintaining competitive completeness and F1 scores,either matching or slightly trailing those of COLMAP. These results underscore the effectiveness of our online incremental mesh generation pipeline, which consistently produces well-balanced, high-quality reconstructions that rival those of state-of-the-art Structure-from-Motion (SfM) systems in overall mesh quality. Fig.~\ref{fig:mesh_gym_roiqualitycompare} provides a qualitative comparison of the reconstructed mesh models across different methods and datasets, with CC serving as the reference. We can see that, although the meshes generated by all three evaluated methods-including Ours-exhibit less geometric detail and surface smoothness compared to those from CC, they successfully capture the dominant structural outlines and key architectural features across all three datasets. Notably, our method generates coarse proxy models that are sufficiently accurate and structurally coherent, making them well-suited for our downstream tasks such as quality assessment and predictive flight path optimization.

\subsubsection{Time efficiency}
\begin{figure}[htbp]
    \centering
    \includegraphics[trim={0 0 0 0},clip, width=0.9\linewidth]{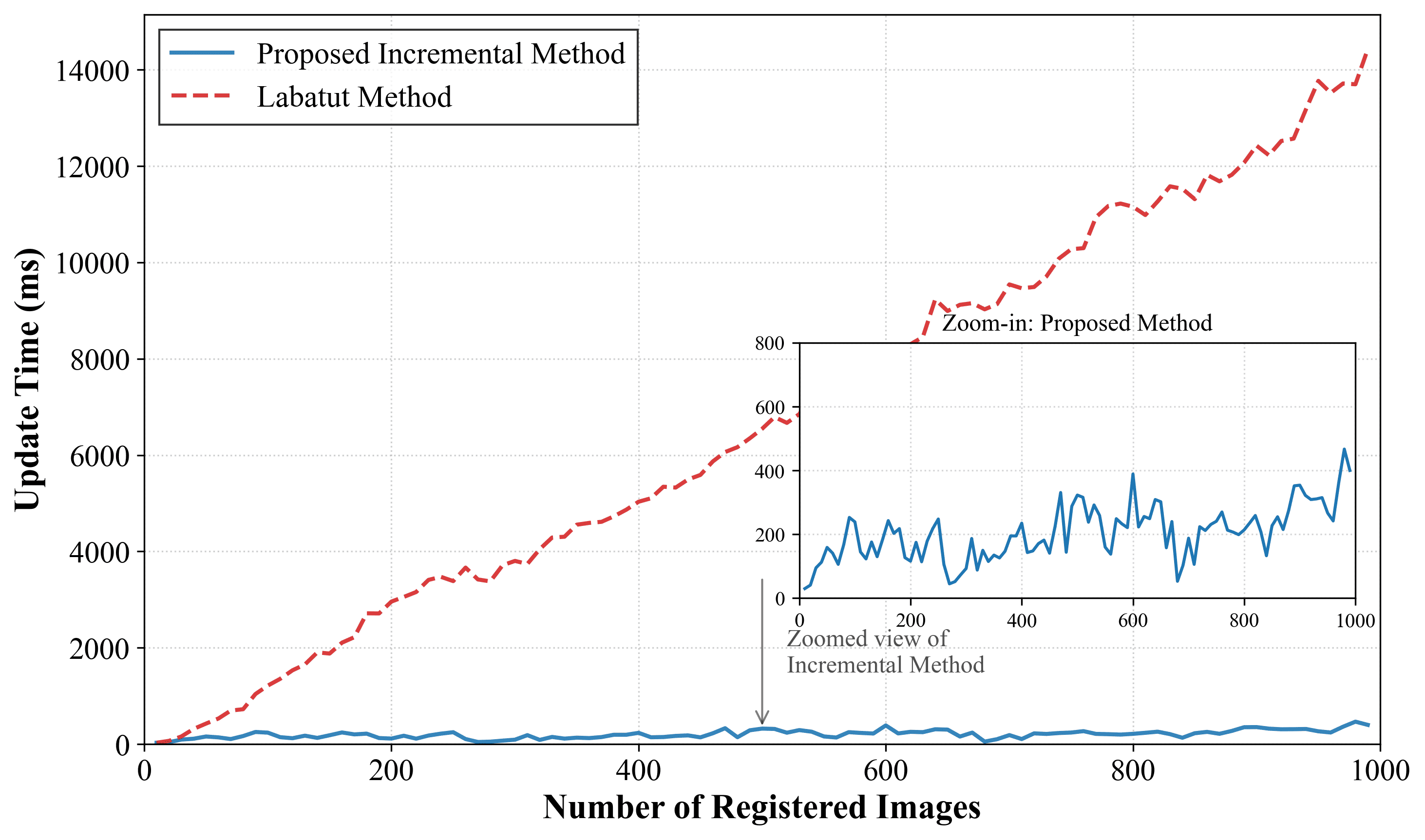}
    \caption{Comparison of cost time with Labatut et al.\cite{labatut2009robust} for updating energy function.}
    \label{fig:mesh_efficiency_com}
     \vspace{-0.25cm}
\end{figure}

\begin{figure}[htbp]
    \centering
    \includegraphics[trim={0 0 0 0},clip, width=0.9\linewidth]{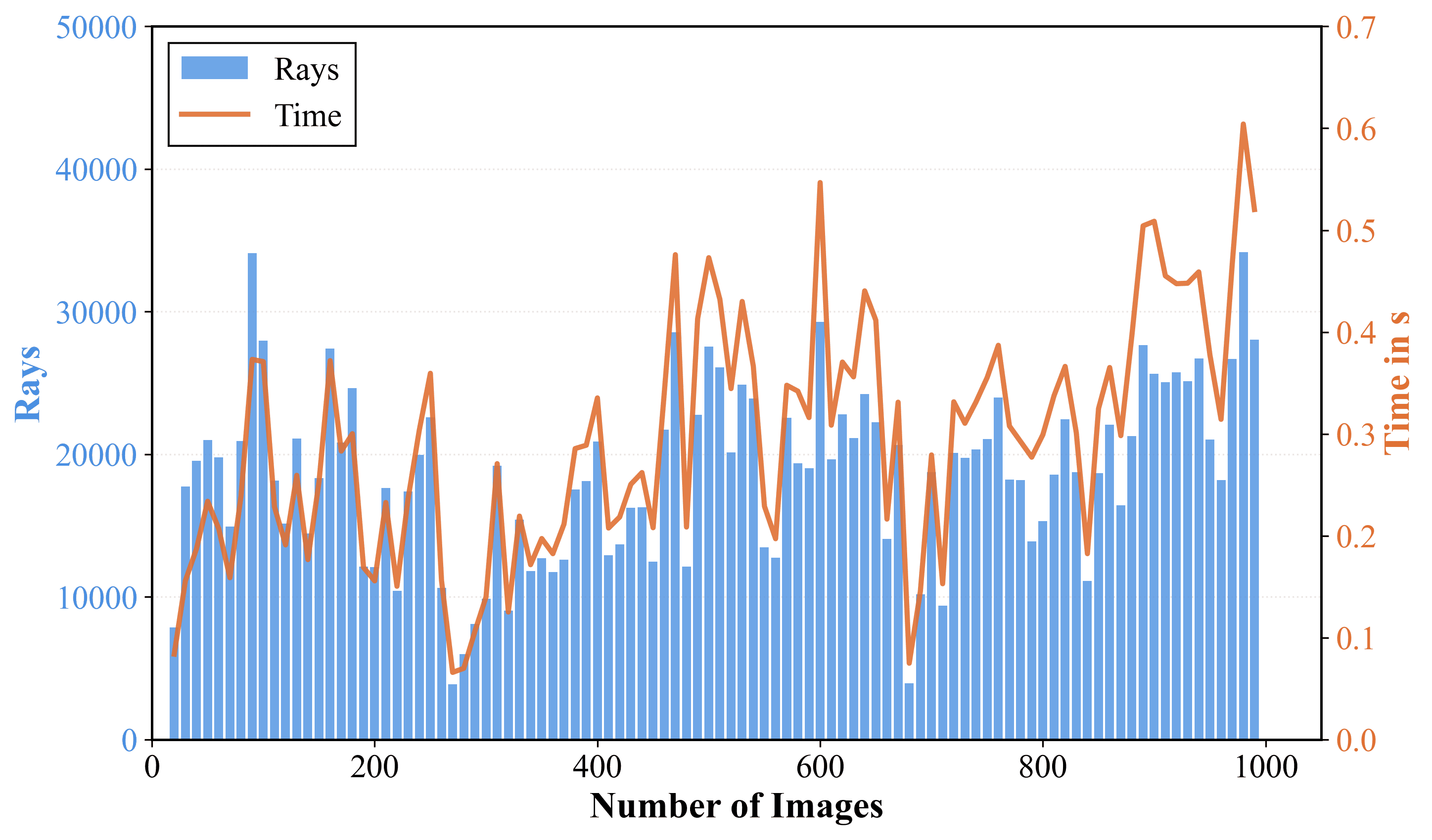}
    \caption{Cost time for energy updating and the number of rays involved during the energy function update.}
    \label{fig:ray_time}
     \vspace{-0.25cm}
\end{figure}

\begin{figure}[htbp]
    \centering
    \includegraphics[trim={0 0 0 0},clip, width=0.9\linewidth]{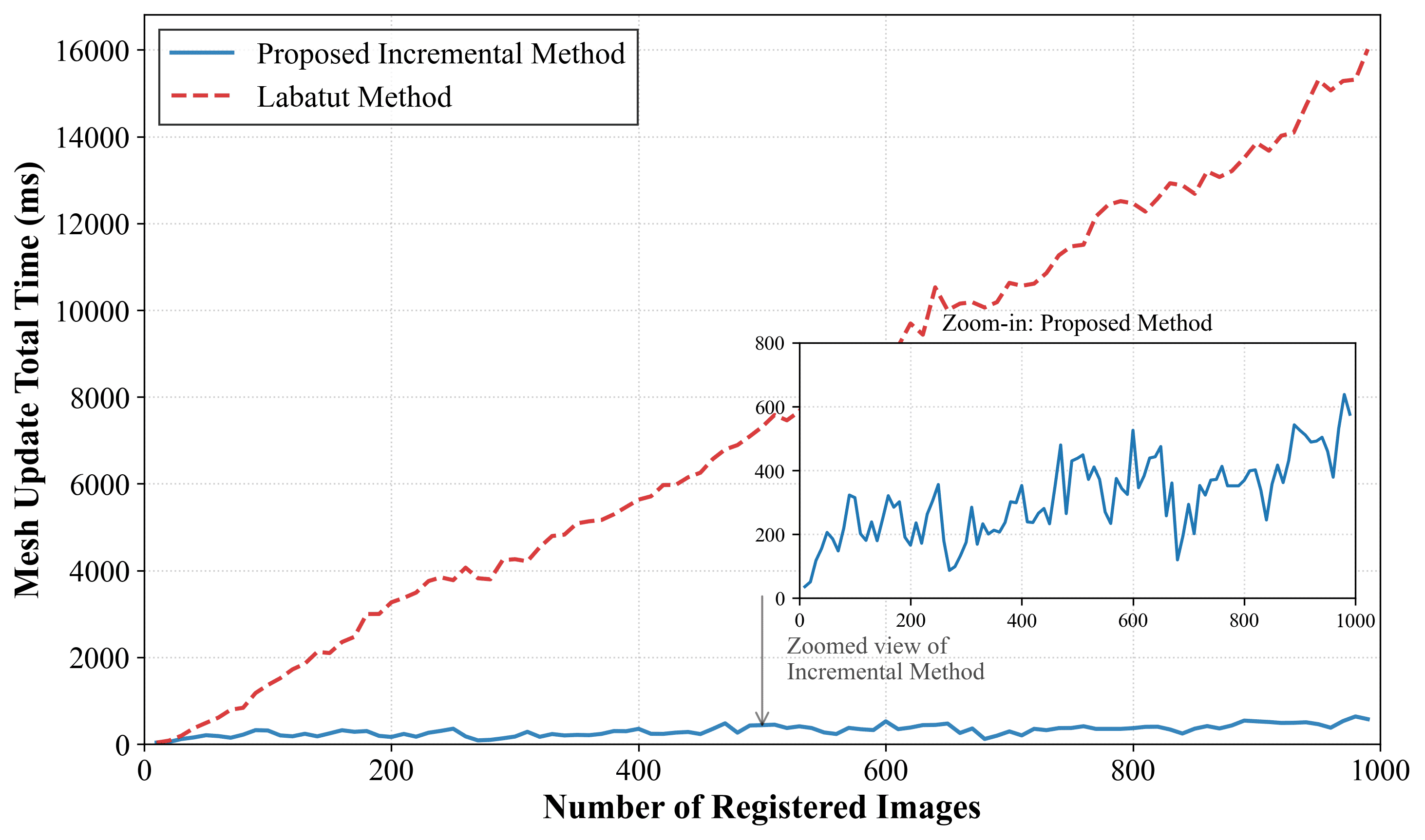}
    \caption{Comparison of cost time with Labatut et al.\cite{labatut2009robust} for incremental mesh updating.}
    \label{fig:mesh_total_comparison}
     \vspace{-0.25cm}
\end{figure}

To analyze the cost time, the \textit{US3D} dataset is used. We conduct a detailed comparative analysis focusing on two key computational stages: the energy function update and the overall incremental mesh updating process. The established method by Labatut et al. \cite{labatut2009robust} is compared.
Fig.~\ref{fig:mesh_efficiency_com} presents the cost time for updating the energy function as the number of registered images increases. It is clearly observed that, while both methods exhibit an increasing in computation time with more images (particularly during the initial stages), our proposed incremental mesh method demonstrates  a significantly slower growth rateand eventually stabilizes as more images are incorporated. In contrast, the baseline Labatut method (red dashed line) shows a steep, near-linear increase, reaching prohibitively high values as the image count approaches 990. Conversely, our approach (solid blue line) maintains consistently low update times throughout, highlighting its superior scalability. The inset plot, which zooms in on the behavior of our method, further reveals that the energy function update time—despite minor fluctuations—remains tightly bounded within a low and stable range, a crucial property for real-time applications. In addition, Fig.~\ref{fig:ray_time} illustrates  the per batch (10 images) processing cost time alongside the number of required rays during energy function updates. A clear positive correlation is observed between these two quantities. Importantly, the number of required rays fluctuates within a bounded range rather than growing monotonically. This bounded behavior is a defining characteristic of our approach, because the energy update is formulated to be local and incremental, it only recomputes rays associated with newly registered or modified regions of the scene. As a result, the per-iteration computational complexity remains inherently constrained. 
As shown in Fig.\ref{fig:mesh_total_comparison}, it depicts the trend of incremental mesh update cost time as the input image are added. The results indicate that when incrementally processing nearly to 1,000 images, our method completes the mesh update in approximately 600 milliseconds, whereas the baseline method requires about 16,000 milliseconds (a more than 26-fold speedup). Notably, the total processing time of our method closely matches the energy optimization time reported in Fig.~\ref{fig:mesh_efficiency_com}, indicating that energy function evaluation constitutes the primary computational bottleneck. Consequently, the efficiency of this component directly governs the overall performance of the proposed pipeline.

\begin{figure}[htbp]
    \centering
    \includegraphics[trim={0 0 0 0},clip, width=0.9\linewidth]{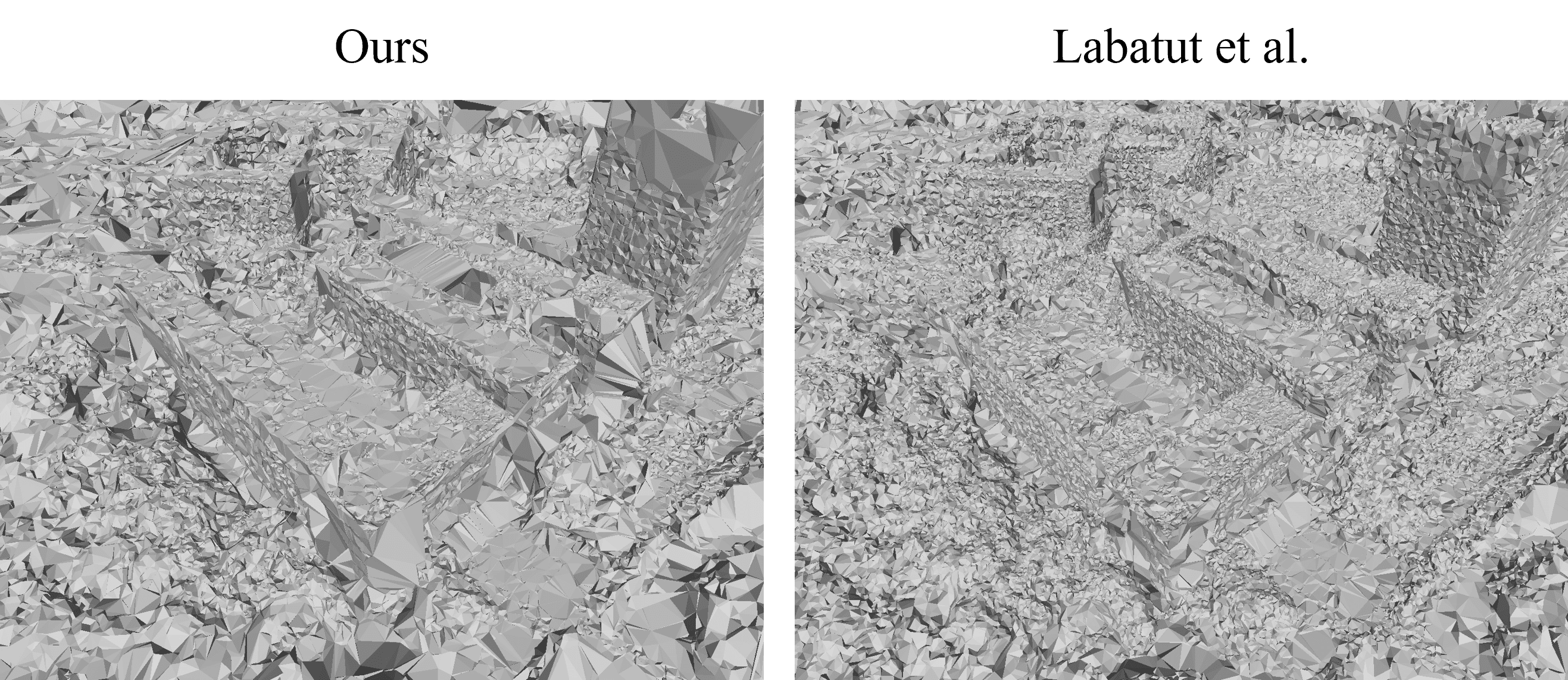}
    \caption{Comparison of mesh results on the US3D dataset. The left image represents the mesh by our method, while the right image shows the result from Labatut et al.\cite{labatut2009robust}.}
    \label{fig:mesh_labatut_quality_comparison}
     \vspace{-0.25cm}
\end{figure}

In Fig.\ref{fig:mesh_labatut_quality_comparison}, a qualitative comparison of the reconstructed mesh on the \textit{US3D} dataset is shown. We can see that, in general, both methods can produce mesh models that coarsely describe the real 3D scene. More specifically, our method appears slightly coarser with larger planar regions, whereas the baseline method yields a relatively more refined and detailed mesh. Despite this difference in refinement, our method still successfully captures the primary geometric structure and major features of the scene, providing a comparable overall representation. This indicates that while our approach slightly sacrifices level of detail, it maintains a fundamentally similar reconstruction quality to the baseline, achieving a balanced trade-off that prioritizes computational efficiency without substantially compromising the geometric integrity of the output.

\label{sec:Experiments}
 \begin{figure*}[!t]  
    \centering
    \includegraphics[width=0.85\textwidth]{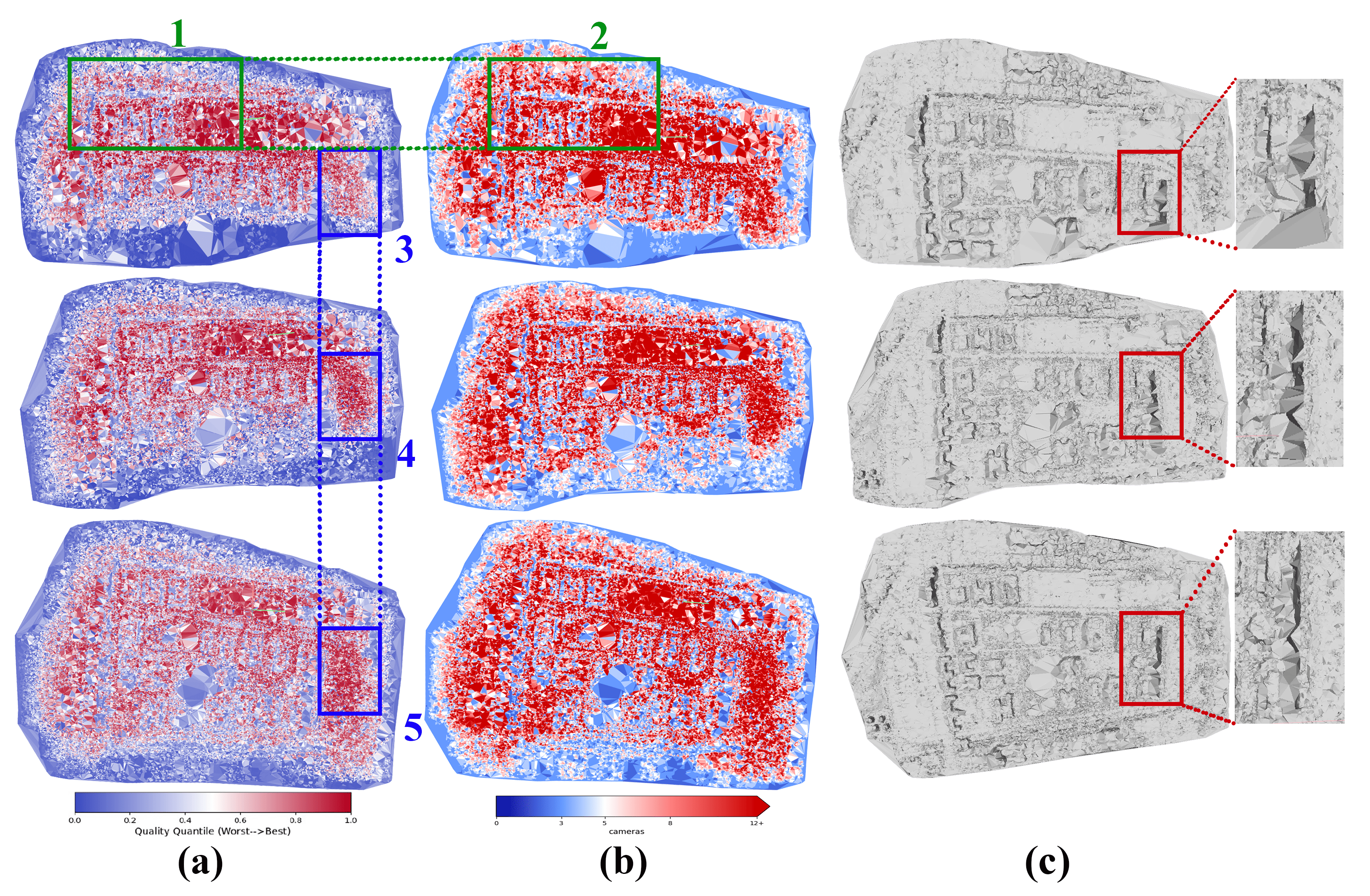}  
    \caption{Online quality assessment of the evolution on the reconstructed scene during image acquisition (the results of the 200th, 300th and 467th captured images are shown in the three rows from top to bottom, respectively). 
    Column (a) visualizes the per-face ensemble quality score $Q_{total}$, where colors range from blue (low reconstruction quality) to red (high reconstruction quality).
Column (b) shows the per-face observation redundancy, representing the number of images observing each surface element—regions in red indicate strong multi-view coverage.
Column (c) displays the reconstructed mesh model with 200, 300 and 467 images for reference using our incremental surface reconstruction.}  
    \label{fig:online_quality_feedback}
\end{figure*}

\begin{figure*}[!t]  
    \centering
    \includegraphics[width=0.95\textwidth]{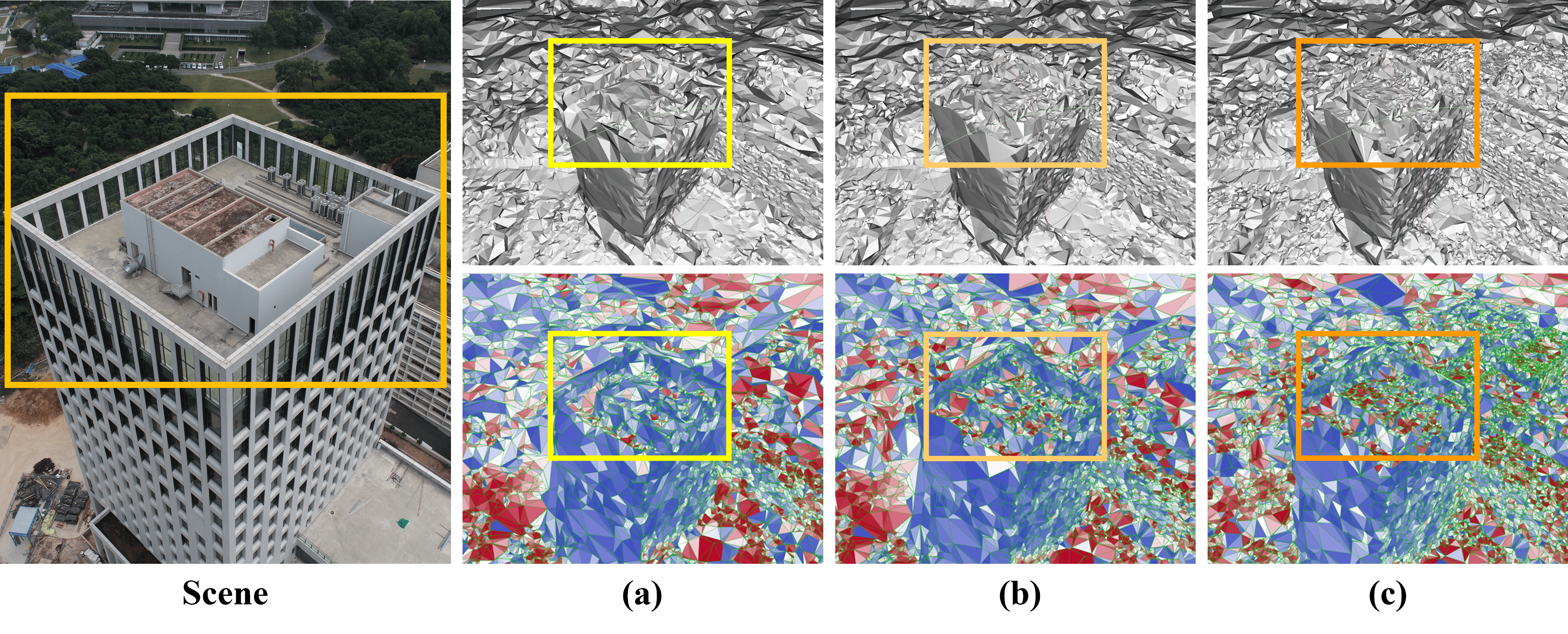}  
    \caption{Localized consistency verification of the quality indicator within a specific Region of Interest (ROI). 
    The “Scene” column presents a sample image of the target building rooftop, with yellow box marking the ROI. Columns (a), (b), and (c) visualize the detailed evolution of the per-face ensemble quality score $Q_{total}$, after integrating 60, 150, and 300 images, respectively. Similar to Fig.~\ref{fig:online_quality_feedback}, the colormap transitions from blue (low quality) to red (high quality). The green line represents the mesh line. This figure illustrates the local evolution of the indicator as image capturing proceeds from sparse to sufficient coverage within local ROI.}  
    \label{fig:quality_roi_assessment}
\end{figure*}

\subsection{Online Quality Assessment}\label{sec:quality_assessment_exp}
To explore how the proposed ensemble indicator $Q_{total}$ reflects reconstruction quality during the online incremental process, two representative datasets are tested: (i) \textit{PHANTOM} is used to analyze temporal and global quality evolution across three acquisition stages—200, 300, and 467 images—as illustrated in Fig.~\ref{fig:online_quality_feedback}, and (ii) a structurally complex rooftop region from \textit{US3D} (Fig.~\ref{fig:quality_roi_assessment}) is applied to validate local geometric consistency in complex scenarios.

\subsubsection{Global validation of the ensemble quality indicator $Q_{total}$}
Across the three acquisition stages (200-, 300-, and 467-images), the spatial distribution of $Q_{total}$ exhibits striking consistency with the per-face observation redundancy. As shown in Fig.~\ref{fig:online_quality_feedback}, regions with dense observation redundancy in Column(b) consistently appear as high-quality areas(red) indicated by $Q_{total}$ in Column(a), while sparsely observed regions simultaneously exhibit low $Q_{total}$ values (blue).  
The zoomed comparisons (green regions Zoom 1–2) further reveal pixel-level agreement between the two indicators: areas marked by weak redundancy—typically resulting from insufficient view overlap—are captured with equally low $Q_{total}$ scores. This coherent correspondence across both global and local scales confirms that $Q_{total}$ reliably reflects both geometric and observational completeness, validating its effectiveness as an online quality indicator.

\subsubsection{Quality evolution during incremental acquisition}
The second set of zoomed regions (blue regions Zoom 3–5 in Fig.~\ref{fig:online_quality_feedback}) illustrates the temporal sensitivity of $Q_{total}$. At earlier stage(Zoom 3), the selected region is predominantly rendered in cooler tones, indicating incomplete geometric constraints and limited observation redundancy. As more images are incorporated(Zoom 4 and Zoom 5), the same region transitions toward warmer tones, reflecting increased observation redundancy and reduced reprojection error. 

In addition to the color-based quality evolution, the corresponding mesh details in Column(c) provide direct visual evidence of structural improvement. The three red-highlighted regions—aligned with Zoom 3–5—show a clear progression in reconstruction quality. Across the three stages, the mesh becomes progressively more coherent and visually stable as more images are integrated, with surfaces exhibiting increasingly detailed and well-structured. This gradual enhancement in mesh regularity closely follows the rise of $Q_{total}$, confirming that the indicator responds faithfully to genuine improvements in reconstructed surface detail.

\subsubsection{Local consistency verification on region of interest}
To further validate the sensitivity of the proposed indicator in geometrically complex areas, we conduct a localized analysis on a rooftop structure containing fine‐scale details, selected as the region of interest (ROI) in Fig.~\ref{fig:quality_roi_assessment} from dataset \textit{US3D}. This structure, characterized by sharp depth discontinuities and narrow layout variations, provides a challenging scenario for evaluating whether $Q_{total}$ can reliably capture incremental improvements in reconstruction quality. Similar to the gloabal validation, for the initial stage (60 input images), the ROI enclosed by the yellow frame is predominantly rendered in cooler tones, indicating limited observation redundancy and unstable geometric constraints within this localized structure. As additional images are incorporated (150 and 300 images), the same ROI exhibits a clear and continuous progression toward warmer tones. The surfaces inside the highlighted region transition from blue to mixed blue–red patterns and ultimately to predominantly red areas, directly reflecting the increased observation redundancy and the reduction in reprojection error.

This localized progression demonstrates that $Q_{total}$ responds consistently to incremental improvements even within small, detail‐rich regions. The behavior is fully aligned with the temporal evolution observed in the global analysis (Fig.~\ref{fig:online_quality_feedback}, Zoom 3–5), confirming that the indicator is capable of capturing both global and fine-grained reconstruction refinement. Such spatially coherent sensitivity is essential for guiding viewpoint allocation in the subsequent online path planning stage.

In summary, the global evolution in Fig.~\ref{fig:online_quality_feedback} and the localized verification in Fig.~\ref{fig:quality_roi_assessment} demonstrate that $Q_{total}$ maintains consistent agreement with observation redundancy, responds predictably to incremental data accumulation, and reliably reflects reconstruction completeness at both scene level and structural-detail level. These properties make the indicator particularly suitable for guiding real-time UAV path planning by accurately highlighting regions that require additional coverage.

\subsection{Path Planning}\label{sec:path_planning_exp}
To validate the performance of our proposed predictive path planning module, we conduct comparison on one of the \textit{US3D} datasets, which provides high-resolution aerial–ground observations and serves as a challenging benchmark for evaluating viewpoint distribution and trajectory quality.

\begin{figure}[htbp]
    \centering
    \includegraphics[trim={0 0 0 0},clip, width=0.90\linewidth]{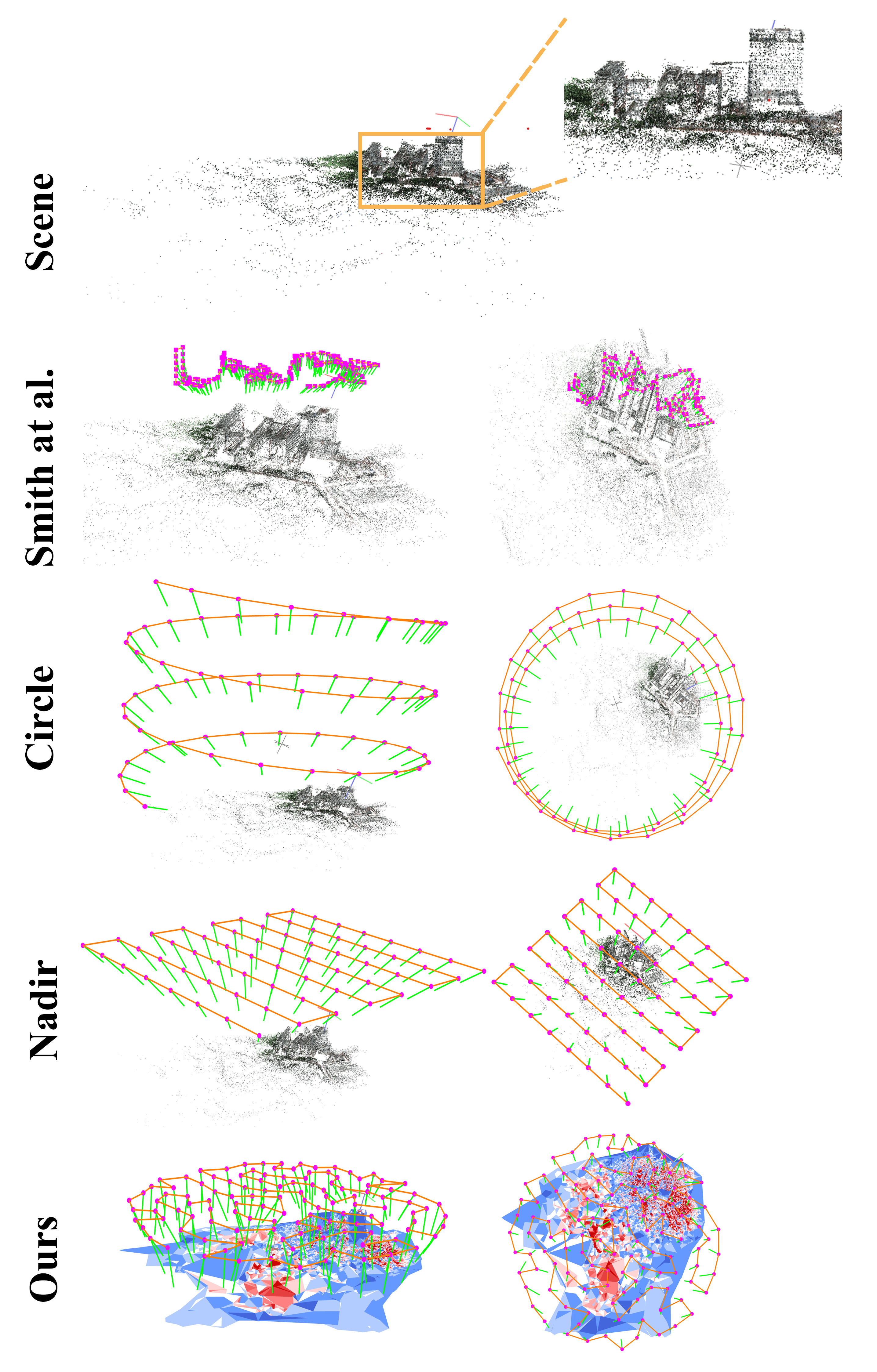}
    \caption{Comparison of various flight planning strategies on the \textit{US3D} dataset. 
    The proposed adaptive path planning is compared with two widely used flight patterns—nadir flight (Nadir) and circling flight (Circle)—as well as a inherent flight path provided in the UrbanScene3D benchmark \cite{UrbanScene3D}. 
    }
    \label{fig:path_exp_compare}
    \vspace{-0.25cm}
\end{figure}

\begin{table}[htbp]
\centering
\caption{Quantitative comparison of four various flight planning strategies on the \textit{US3D} dataset.}
\label{tab:quantitative_path_compare}
\vspace{-0.2cm}
\renewcommand{\arraystretch}{1.2}

\begin{tabular}{cccc}
\hline
Method & VP Num & TLength & GTime(ms) \\
\hline
\textit{Smith et al.}~\cite{smith2018aerial} & 230 & 387.86 & /(Offline) \\
\textit{Circle} & 66 & 1867.03 & /(Manual Designed)\\
\textit{Nadir} & 64 & 1783.99 &/(Manual Designed)\\
\textit{Ours} & 120 & 947.43  & 744.342\\
\hline
\end{tabular}
\vspace{-0.3cm}
\end{table}

\subsubsection{Baseline Flight Strategies for Comparison}
we compare the proposed mesh quality-driven adaptive flight strategy against two commonly designed UAV paths: 

(i) \textbf{Nadir flight path (\textit{Nadir})}: 
A grid-based flight pattern executed at a constant altitude with fixed nadir viewing direction ($90^\circ$ downward).  
The flight altitude $h_{\text{nadir}}$ is set to $0.5R$, where $R$ is the scene bounding sphere radius, 
ensuring sufficient distance above the scene. 
The grid spacing $d_{\text{grid}}$ is configured as $0.2R$ to maintain uniform coverage, 
with the coverage area extending to $0.8R$. 

(ii) \textbf{Circling path (\textit{Circle})}: 
A helical trajectory flied around the target structure with look-at-center orientation.  
The circling radius $r_{\text{circle}}$ is set to $0.7R$, 
with the flight altitude varying from $h_{\min} = 0.2R$ (base height) 
to $h_{\max} = 1.2R$ (top height), spanning a vertical range of $1.0R$. 
The trajectory comprises 3 helical layers with angular sampling interval $\Delta\theta = 20^\circ$.

In practice, these two paths are usually preferred to balance simplicity and scene coverage. As illustrated in Fig.~\ref{fig:path_exp_compare}, two baselines (\textit{Nadir} and \textit{Circle}) generate sparse viewpoints to cover the scene. In contrast, the proposed flight path planning reallocates viewpoints primarily toward the detected low-quality regions(blue), demonstrating its ability to guide the UAVs effectively “explore” the scene and then “exploit” the identified areas through planned image capturing.

In addition to the two baselines, we include the UrbanScene3D inherent trajectory (denoted as “\textit{Smith et al.}”).
This trajectory is generated using the continuous optimization method of Smith et al.~\cite{smith2018aerial} and is also officially released as part of the UrbanScene3D benchmark\cite{UrbanScene3D}.
Unlike manually designed flight paths, this trajectory is generated by optimizing viewpoint placement over a complete coarse proxy model to maximize expected multi-view stereo performance. Fig.~\ref{fig:path_exp_compare} shows that \textit{Smith et al.} trajectory places most of its viewpoints around the dominant building structures, which occupy the enlarged regions of the \textit{Scene}. However, compared with \textit{US3D}, our strategy distributes viewpoints across the entire \textit{Scene}, with a clear emphasis on the identified low-quality regions(blue) using the proposed mesh-quality indicator, enabling more balanced coverage and ultimately contributing to improved global reconstruction quality.

\subsubsection{Quantitative Comparison}
To further complement the qualitative comparison in Fig.~\ref{fig:path_exp_compare}, Tab.~\ref{tab:quantitative_path_compare} presents a quantitative comparison of the four flights on the \textit{US3D} dataset.
Each method is assessed using three metrics that directly reflect operational performance in real UAV deployments: 
(i) the number of selected viewpoints(denoted as "VP Num"), indicating the effort spent on capturing images, and 
(ii) the total trajectory length(denoted as "TLength"), which is closely tied to flight time, battery consumption, and thus overall mission cost, and
(iii) the trajectory-generation time(denoted as "GTime"), which reveals the responsiveness of the path planning module and its compatibility for real-time operation.

As shown in Tab.~\ref{tab:quantitative_path_compare}, the proposed flight strategy reallocates a larger number of viewpoints while simultaneously achieving a shorter overall trajectory length compared with the two very practical baselines (\textit{Nadir} and \textit{Circle}), thereby reducing the flight effort.
When comparing to \textit{Smith et al.}\cite{smith2018aerial}, our method yields fewer images but results in a longer flight path.  This extended flight path-almost 3 times longer-enables the UAVs to better "explore" the entire \textit{Scene}, rather than concentrating observations primarily around a single structure.
Moreover, the trajectory is generated within a fraction of a second, demonstrating that the planning module is fast enough to operate within an online reconstruction process.

\subsection{Performance of Online Explore-and-Exploit Feedback Pipeline}\label{sec:entire_pipeline_exp}
In this section, to evaluate the effectiveness and practicality of the proposed On-the-fly Feedback SfM framework in a realistic online data acquisition setting, where reconstruction quality continuously guides subsequent viewpoint generation and in turn influences the evolving reconstruction quality, we conduct an experiment on the \textit{XingHu} dataset, a real large-scale outdoor scene captured with high-resolution UAV imagery.

\label{sec:Experiments}
 \begin{figure*}[!t]  
    \centering
    \includegraphics[width=0.95\textwidth]{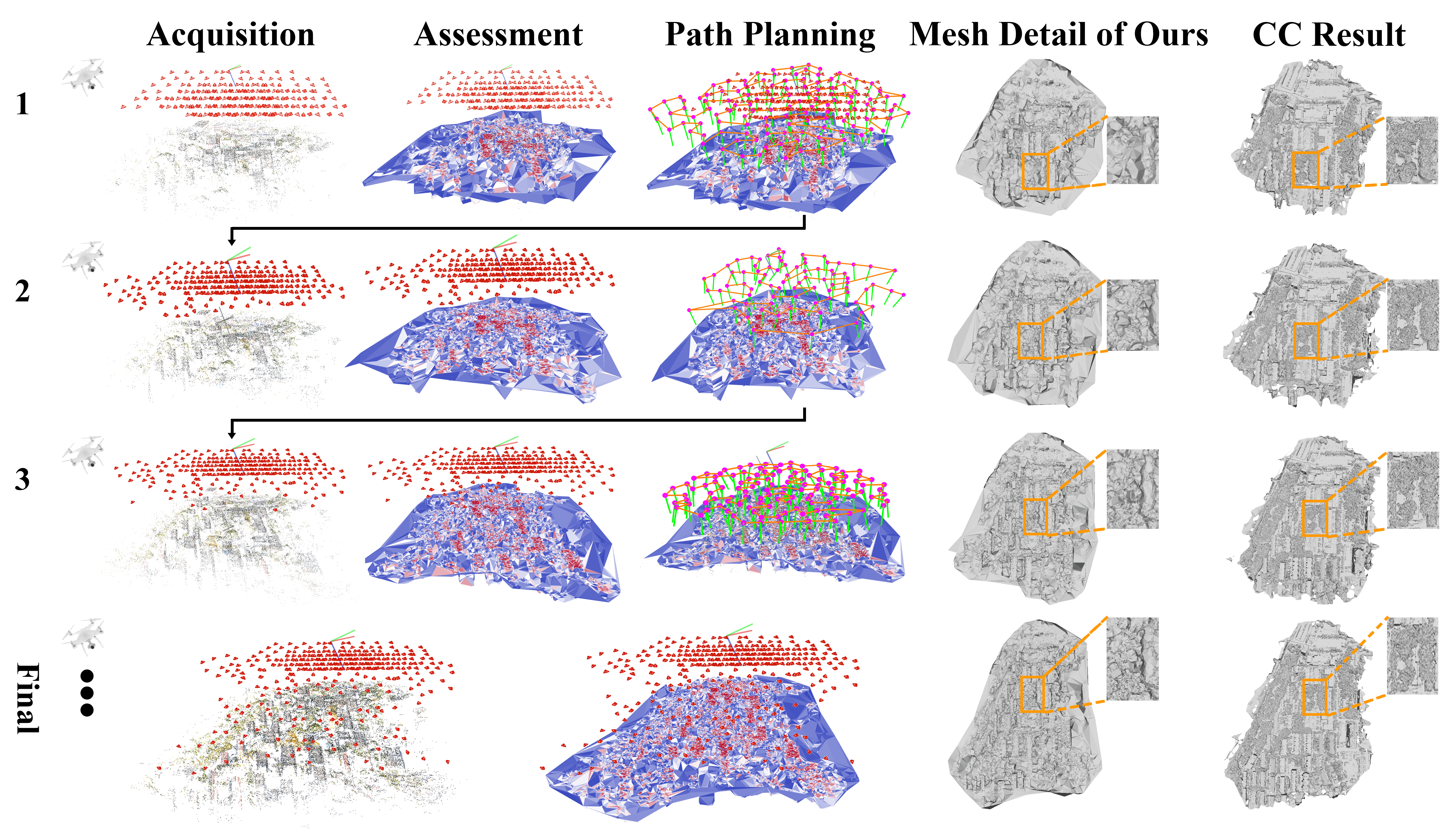}  
    \caption{Visualization of the proposed explore-and-exploit online feedback framework during a realistic online acquisition task of \textit{Xinghu} dataset. The rows, from top to bottom, illustrate the temporal evolution of the reconstruction process. The columns display the intermediate states: sparse point cloud acquisition, geometric feedback, adaptive path planning, and local mesh details. The final row presents the completed reconstruction of the target scene, and the last column is the reconstructed fine-grained mesh model by using CC with the corresponding captured images.}  
    \label{fig:explore_and_exploit}
\end{figure*}

\begin{table*}[htbp]
\centering
\caption{The quantitative results and time efficiency during online incremental reconstruction.}
\label{tab:explore_and_exploit}
\vspace{-0.2cm}
\renewcommand{\arraystretch}{1.2}

\begin{tabular}{cccccc>{\centering\arraybackslash}m{3cm}}
\hline
\makecell{Row} & \makecell{Accuracy} & \makecell{Completeness} & \makecell{F1 Score} 
& \makecell{Average time per Image \\ (in second)} 
& \makecell{Trajectory Generation \\ time (in ms)} \\
\hline

1     & 0.6764 & 0.4143 & 0.5138 & 1.465 & 602.311  \\
2  & 0.8333 & 0.5232 & 0.6428 & 1.470 & 183.801  \\
3    & 0.8141 & 0.4830  & 0.6063 & 1.415 & 497.513  \\
Final    & 0.8291  & 0.5099 & 0.6315 & 1.475 & /  \\
\hline
\end{tabular}
\vspace{-0.3cm}
\end{table*}

The entire pipeline is initiated when the UAV flies to the target area-\textit{XingHu} and captures an initial set of images, which are then processed online to generate the sparse point clouds and camera poses. Subsequently, online incremental surface reconstruction and mesh quality assessment are performed based on the newly generated point cloud. Leveraging this mesh-quality driven feedback, an updated flight path is computed in real time. The UAV then executes this optimized trajectory—“exploiting” the newly planned path—to capture additional imagery, thereby progressively expanding the explored region while simultaneously enhancing reconstruction fidelity. As the UAV continuously captures images in-flight, the full explore-and-exploit loop, comprising image acquisition, reconstruction, quality assessment, flight path planning, and re-acquisition, operates incrementally and in near real time. This iterative process continues until a final, high-quality 3D reconstruction is achieved. Within iteration, the coarse mesh model of the scene is progressively extended and refined, yielding reconstruction of increasing completeness, geometric detail and accuracy. Fig.~\ref{fig:explore_and_exploit} visually exhibits the evolution of the reconstruction across multiple online feedback iterations, including the final stage. Specifically, the first column (\textit{Acquisition}) displays the online-generated sparse point cloud and camera poses; the second and fourth columns present the incremental coarse surface model and its associated mesh-quality assessment, respectively; the third column shows the predictive flight paths generated in response to the current reconstruction quality; and the final column depicts reference mesh models produced by CC, using only the images captured up to that iteration as input.

As shown by the \textit{Mesh Detail of Ours} of Fig.~\ref{fig:explore_and_exploit}, the orange-boxed regions reveal progressively finer and more well-detailed geometric structures as reconstruction evolves. Furthermore, when comparing the results of 1th row to 3rd row, we can see that the proposed framework can effectively generate more complete reconstruction of the scene during the online processing.

To quantitatively evaluate our performance, as it is shown in Tab.~\ref{tab:explore_and_exploit}, all three evaluation metrics exhibit an obvious increasing trend as the explore–and-exploit iterations proceed. From Row 1 to Row 3, both accuracy and F1 score show steady improvement, indicating that the reconstructed geometry becomes increasingly precise while maintaining a balanced trade-off between accuracy and completeness. This increasing behavior in quality metrics suggests that our incremental acquisition and online mesh-quality driven feedback mechanism can steadily enhance the overall quality of the evolving mesh.
The final reconstructed mesh achieves an accuracy of 0.8291 and an F1 score of 0.6315, reflecting a well-balanced trade-off between geometric fidelity and structural completeness at the end of the incremental reconstruction process. Although the completeness (0.5099) is marginally lower than the value achieved in Row 2, the overall performance, evaluated under the precision–recall framework, demonstrates that the final model retains consistent quality and structural reliability.

Furthermore, the average processing time per image remains stable (around 1.4–1.5 seconds) throughout the test, indicating that the proposed incremental processing pipeline  sustains near real-time performance across all iterations. Notably, trajectory generation time remains consistently below one second, enabling the feedback-driven system to rapidly incorporate newly captured images and promptly update the UAV’s flight trajectory during the “explore”. 

In summary, the consistent qualitative evolution of the reconstruction alongside stable quantitative performance underscores the robustness and efficacy of our framework in real-world, large-scale scenarios. By tightly coupling image acquisition, online reconstruction quality assessment, and adaptive path planning, our method continuously identifies poorly reconstructed regions and dynamically generates updated flight trajectories in real time. This closed-loop capability empowers UAVs to autonomously navigate along optimized paths to acquire supplementary imagery, effectively mitigating coverage gaps commonly associated with pre-planned flight missions. Consequently, the need for redundant post-mission re-flights is eliminated, ensuring both high acquisition efficiency and enhanced reconstruction completeness.

\section{Conclusion}
In this paper, we introduce On-the-fly Feedback SfM, an online \textit{explore-and-exploit} UAV photogrammetry framework that enables in-situ reconstruction and adaptive mesh quality-driven flight path optimization. In particular, built upon \textit{on-the-fly SfM}, our method seamlessly integrates three core components: (i) an incremental surface reconstruction method based on dynamic graph cuts for rapid surface extraction; (ii) a mesh-aware quality assessment mechanism that fuses multi-view geometric cues to quantify local reconstruction fidelity;(iii) a predictive path planning strategy that leverages quality feedback to optimize UAV trajectories in real time. Extensive evaluations on both self-captured datasets and public benchmarks demonstrate that, our method can achieve a superior trade-off between reconstruction efficiency and model quality compared to conventional approaches. Crucially, the proposed framework dynamically explores the scene during flight, identifying under-reconstructed or low-quality regions on-the-fly, thereby significantly reducing coverage gaps and eliminating the need for costly post-mission re-flights.

Together, these proposed components form a closed-loop acquisition–reconstruction–feedback–planning–acquisition cycle, offering a robust, adaptive, and real-time-capable solution for time-critical applications such as rapid digital twin generation and emergency response scenarios where both speed and geometric reliability are paramount.



\bibliographystyle{IEEEtran.bst}
\bibliography{references}

\end{document}